\documentclass[10pt]{article}

\usepackage{PRIMEarxiv}
\usepackage[T1]{fontenc}
\usepackage{times}
\usepackage{microtype}
\usepackage{xcolor}
\usepackage[hyphens]{url}
\usepackage{graphicx}
\usepackage{hyperref}
\usepackage{caption}
\usepackage{wrapfig}
\usepackage{booktabs}
\usepackage{multirow}
\usepackage{makecell}
\usepackage{colortbl}
\usepackage{amsmath}
\usepackage{amssymb}
\usepackage{enumitem}
\usepackage[capitalize,noabbrev]{cleveref}

\usepackage{amsmath,amsfonts,bm}
\usepackage{xspace}

\def\eqref#1{equation~\ref{#1}}

\def\1{\bm{1}}

\def\rmC{{\mathbf{C}}}

\def\rmI{{\mathbf{I}}}

\def\rmR{{\mathbf{R}}}

\def\vmu{{\bm{\mu}}}

\def\ve{{\bm{e}}}

\def\vp{{\bm{p}}}

\def\vs{{\bm{s}}}

\def\vv{{\bm{v}}}

\def\vx{{\bm{x}}}

\def\vz{{\bm{z}}}

\DeclareMathAlphabet{\mathsfit}{\encodingdefault}{\sfdefault}{m}{sl}
\SetMathAlphabet{\mathsfit}{bold}{\encodingdefault}{\sfdefault}{bx}{n}

\def\gD{{\mathcal{D}}}

\def\gT{{\mathcal{T}}}

\def\sR{{\mathbb{R}}}

\DeclareMathOperator*{\argmax}{arg\,max}

\makeatletter
\DeclareRobustCommand\onedot{\futurelet\@let@token\@onedot}
\def\@onedot{\ifx\@let@token.\else.\null\fi\xspace}

\def\ie{\emph{i.e}\onedot}

 \def\vs{\emph{vs}\onedot}

\makeatother

\newcommand{\gen}[1]{\langle #1 \rangle}
\title{EEG Decoding in Context: A Framework for Tuning-Free EEG Foundation Models}

\author{Hao-Run Cai$^{1}$ \quad Si-Yang Liu$^{1}$ \quad Han-Jia Ye$^{1}$ \quad Hairong Zheng$^{2}$\\
  $^{1}$National Key Laboratory for Novel Software Technology, Nanjing University\\
  $^{2}$Institute of Brain-Machine Interface, Nanjing University\\
  \texttt{\{caihr,liusy,yehj\}@lamda.nju.edu.cn} \quad \texttt{hr.zheng@nju.edu.cn}}
\date{}

\begin{document}
\maketitle
\begingroup
\renewcommand{\thefootnote}{}
\makeatletter
\let\@footnotetext\H@@footnotetext
\makeatother
\endgroup
\begin{abstract}
Electroencephalography (EEG) decoding is a central route toward brain--computer interfaces for communication, rehabilitation, cognitive assessment, and neurological monitoring. Recent EEG foundation models have substantially advanced reusable signal representations, but their downstream use still typically requires a task-specific decoder, gradient-based adaptation, and validation-driven model selection. We investigate a complementary foundation-model paradigm based on in-context learning: labeled EEG recordings define the decoding task as context, and new recordings are predicted without updating model parameters. We first pair raw EEG features with a pretrained general-purpose in-context predictor. This direct formulation provides a promising starting point, but its performance remains limited and unstable and does not generalize reliably across EEG paradigms. We find that suitable context construction can address these limitations by improving the effective signal-to-noise ratio and exposing temporal and spatial structure. Building on this insight, we introduce a general framework that combines EEG-specific context construction with a frozen in-context predictor and a metric-based context adaptation mechanism, which matches each downstream task to an appropriate context representation without relying on validation data. Across ten datasets spanning seven EEG decoding paradigms, our method achieves the highest mean BAcc among eleven specialized and fine-tuned EEG baselines, without task-specific parameter updates or validation-based model selection. This framework provides a practical path toward tuning-free EEG foundation models in which adaptation is expressed through context rather than parameter updates.
\end{abstract}

\section{Introduction}
\label{sec:introduction}

\begin{wrapfigure}{r}{.55\textwidth}
  \centering
  \vspace{-10pt}
  \includegraphics[width=\linewidth]{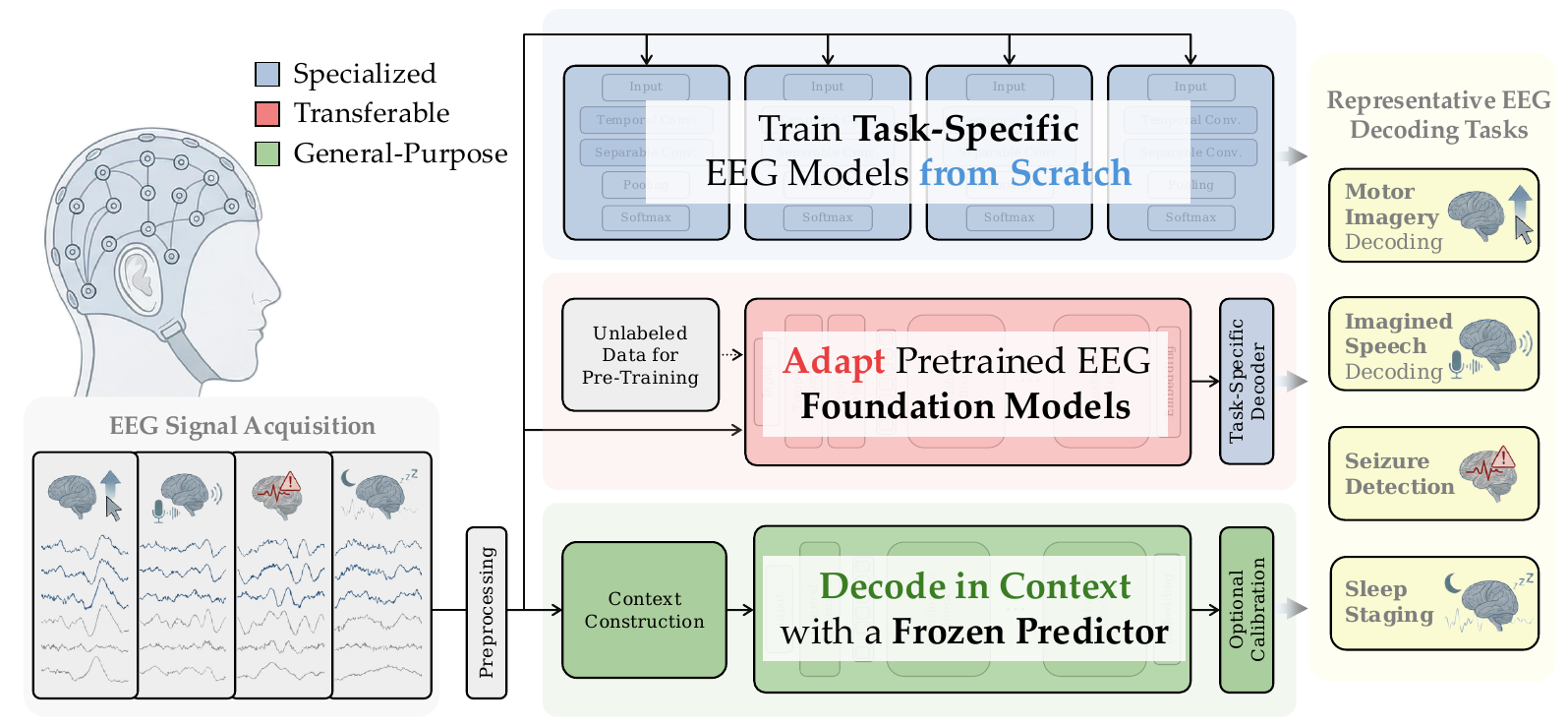}
  \caption{\textbf{Three routes to EEG decoding.} Specialized models are trained separately for each task; representation-transfer methods reuse a pretrained EEG encoder but still optimize task-specific parameters; our framework instead conditions a frozen ICL predictor on labeled EEG context, expressing task adaptation through context construction.}
  \label{fig:intro_comparison}
  \vspace{-15pt}
\end{wrapfigure}

Electroencephalography (EEG) provides a portable and comparatively low-cost means of recording brain electrical activity at high temporal resolution. Decoding EEG has consequently become central to brain--computer interfaces (BCIs) and related neurotechnology applications, including assistive communication and control, neurorehabilitation, cognitive-state assessment, sleep analysis, affective computing, and neurological monitoring \cite{lotte2018review,saeidi2021neuraldecoding,chevallier2024moabb}. Progress in signal processing and deep learning has enabled increasingly capable decoders across motor imagery, event-related potentials, steady-state responses, sleep, seizure, and clinical paradigms \cite{schirrmeister2017deeplearning,lawhern2018eegnet,banville2026neuralbench}. These advances make reusable EEG decoding models scientifically and practically important: a useful model should transfer accumulated knowledge to a new decoding problem rather than require training anew for every task.

Deployment remains challenging because EEG is low-amplitude, artifact-prone, and nonstationary, while its distribution varies across subjects, sessions, devices, and recording conditions \cite{saha2020variability,wu2022transfer}. Across paradigms, class-discriminative information appears in distinct temporal, spectral, and spatial structures \cite{ang2012filterbank,congedo2017riemannian,lawhern2018eegnet,zhou2025csbrain}. A practical decoder must therefore adapt from limited labeled recordings without assuming that one representation or decision rule transfers unchanged. Calibration can itself be lengthy and user dependent \cite{huang2021calibration}, and collecting more trials from the same small participant pool does not necessarily substitute for participant diversity \cite{bomatter2025diversity}. This creates a recurring requirement: use the labeled EEG that is available for a task to predict held-out recordings quickly, reliably, and with little target-time optimization.

Recent EEG foundation models pursue reusability primarily through pretrained encoders. BIOT introduces frequency-aware tokenization for heterogeneous biosignals \cite{yang2023biot}. LaBraM scales masked pretraining over large EEG corpora \cite{jiang2024labram}. CSBrain explicitly models cross-scale temporal and regional dependencies \cite{zhou2025csbrain}. Although their downstream protocols differ, applying these models to a new EEG task generally requires learning at least a task-specific prediction component; many methods further optimize the backbone or lightweight modules and use validation data to select checkpoints or protocols, incurring additional computation and tuning expertise \cite{lee2025capable}. Pretraining therefore provides reusable representations, but does not usually yield a complete decoder that can be deployed unchanged. As illustrated in \cref{fig:intro_comparison}, specialized training and representation transfer both adapt to a new task through parameter optimization; we investigate a third route in which \emph{labeled context specifies the task}.

In-context learning (ICL) offers a complementary adaptation interface: a predictor receives labeled examples and a query together and produces a prediction while its pretrained parameters remain fixed \cite{brown2020language}. General-purpose ICL predictors such as TabPFN \cite{hollmann2022tabpfn,hollmann2025accurate} and TabICL \cite{qu2025tabicl} demonstrate this capability across heterogeneous structured prediction problems. As a first attempt, we vectorize each EEG trial and provide the raw features directly to a frozen ICL predictor. This simple formulation is promising on several tasks, but its performance is uneven across paradigms and not consistently competitive. Our results point to context construction as a central design challenge: raw vectorization preserves the waveform, yet may foreground nuisance variation and obscure the spectral, spatial, and local temporal relationships on which different EEG tasks depend. We therefore ask how EEG-specific context construction can reduce nuisance variation and expose task-relevant structure to a frozen predictor.

To answer this question, we introduce a general framework that combines EEG-specific context construction, a frozen ICL predictor, a support-only context adaptation rule, and optional post-hoc calibration. Our contributions are:
\begin{itemize}[leftmargin=1.2em,itemsep=2pt,parsep=0pt,topsep=3pt]
  \item We introduce a general framework for constructing tuning-free EEG foundation models from frozen general-purpose in-context predictors, in which labeled EEG context replaces task-specific parameter adaptation at deployment.
  \item We identify context construction as the key interface between EEG and a frozen predictor, and develop complementary waveform, spectral, covariance, regional, and local-temporal contexts together with a metric-based adaptation rule that selects a candidate context using labeled support data alone, without validation performance or query information.
  \item Across ten datasets spanning seven EEG decoding paradigms, our support-only Metric route achieves the highest mean BAcc among eleven specialized and fine-tuned EEG baselines, improves over raw context on every dataset, without downstream gradient optimization or validation-based selection.
\end{itemize}

\section{Related Work}
\label{sec:related_work}

\subsection{Specialized EEG Decoding}

EEG decoding has traditionally been developed through task-specific signal-processing and prediction pipelines. Classical approaches combine temporal or frequency filtering with spatial filters and shallow classifiers \cite{lotte2018review}. Filter-bank common spatial patterns expose band-specific discriminative activity for motor imagery \cite{ang2012filterbank}, while covariance-based methods represent cross-channel relationships on the symmetric positive-definite manifold \cite{barachant2012multiclass,congedo2017riemannian}. These approaches encode explicit assumptions about where class information lies, yet large-scale evaluations show that their relative performance depends on the dataset and paradigm, with covariance-based pipelines remaining competitive in many BCI settings \cite{chevallier2024moabb}.\looseness=-1

Deep specialized decoders replace parts of this handcrafted pipeline with learned temporal and spatial operations. Convolutional architectures learn directly from EEG waveforms \cite{schirrmeister2017deeplearning}, and EEGNet uses compact temporal and depthwise spatial filters across several BCI paradigms \cite{lawhern2018eegnet}. Subsequent architectures explore convolutional, multibranch, and transformer-based temporal--spatial modeling \cite{song2021transformer,peh2022cnntransformer,li2022ffcl,song2023eegconformer,jing2023sparcnet,yang2023contrawr}. Transfer learning, alignment, semi-supervised learning, and calibration-reduction methods further address subject and session shift, but still fit or select task-specific components using target data \cite{wu2022transfer,huang2021calibration}. Collectively, these methods show that useful EEG evidence can be captured by waveform, spectral, covariance, spatial, or local temporal representations. These established signal axes motivate both our complementary context families and the support-set descriptors used to adapt context construction across tasks.

\subsection{EEG Foundation Models}

EEG foundation models seek to replace task-by-task representation learning with reusable pretrained encoders. Models such as BIOT, LaBraM, EEGPT, and CBraMod explore cross-dataset or large-scale self-supervised pretraining over heterogeneous EEG signals \cite{yang2023biot,jiang2024labram,wang2024eegpt,wang2025cbramod}. Subsequent work extends reusable representation learning across modalities, acquisition setups, anatomical regions, temporal scales, and electrode topologies \cite{xiao2025brainomni,elouahidi2025reve,doner2025luna,zhou2025csbrain,fang2025neuript}. Across these designs, the pretrained encoder remains the principal reusable object.

Downstream deployment, however, still commonly specifies a new task through optimized parameters. Depending on the model and protocol, this may involve full fine-tuning, parameter-efficient adaptation, or learning a prediction head or aggregation module. Comparative evidence shows that the selected adaptation protocol materially changes both accuracy and resource use \cite{lee2025capable}. NeuralBench further motivates evaluation across heterogeneous tasks rather than conclusions drawn from a single paradigm \cite{banville2026neuralbench}. EEG foundation models therefore make representations reusable, but generally retain parameter optimization as the interface for task adaptation. In-context learning offers a complementary route in which labeled examples specify the task directly.

\subsection{In-Context Learning}

In-context learning conditions a pretrained predictor on labeled examples and a query, producing a task-specific prediction without updating the predictor's parameters \cite{brown2020language}. General-purpose ICL predictors provide a concrete instance of this interface. TabPFN pretrains a transformer over synthetic prediction problems and conditions predictions on labeled rows from a new dataset \cite{hollmann2022tabpfn,hollmann2025accurate}; TabICL extends this principle to larger structured datasets \cite{qu2025tabicl}. Because these predictors are pretrained independently of EEG, they provide a useful basis for asking whether EEG task adaptation can be expressed through a context interface rather than EEG-specific parameter updates.

ECHO provides an EEG-specific approach to contextual decoding \cite{liu2026echo}. It learns contextual behavior through a meta-learning-like support--query objective and autoregressively predicts task and label tokens without inference-time parameter updates. Its held-out evaluations cover new datasets only within EEG task families represented during training. Despite avoiding parameter updates, ECHO remains computationally inefficient at inference because each target trial requires a separate support-conditioned autoregressive decoding pass. Our framework instead reuses a general-purpose ICL predictor pretrained independently of EEG and makes EEG-specific context construction the task-adaptation interface.

\section{EEG Decoding as In-Context Prediction}
\label{sec:framework}

\begin{wraptable}{r}{.59\textwidth}
  \vspace{-13.5pt}
  \centering
  \small
  \caption{\textbf{Frozen ICL is promising but representation-sensitive.} BAcc of the same frozen ICL predictor under three EEG context representations. Dummy ($1/K$), EEGNet, and CSBrain are references. Each cell is colored according to the strongest baseline it outperforms. \textcolor[HTML]{806000}{\bf Yellow} denotes Dummy, \textcolor[HTML]{006B3C}{\bf green} denotes EEGNet, and \textcolor[HTML]{5E2B97}{\bf purple} denotes CSBrain. \underline{underlining} marks the best ICL performance.}
  \label{tab:raw_probe}
  \setlength{\tabcolsep}{5pt}
  \begin{tabular}{lcccccc}
    \toprule
    \multirow{2.5}{*}{Dataset}
      & \multirow{2.5}{*}{\textcolor[HTML]{806000}{\bf Dummy}}
      & \multirow{2.5}{*}{\textcolor[HTML]{006B3C}{\bf EEGNet}}
      & \multirow{2.5}{*}{\textcolor[HTML]{5E2B97}{\bf CSBrain}}
      & \multicolumn{3}{c}{ICL Predictor} \\
    \cmidrule(lr){5-7}
      & & & & Raw & Spec & Riem \\
    \midrule
    BCIC2020-3
      & .2000
      & .2740
      & .6013
      & \cellcolor[HTML]{C8F7D2}\underline{.4493}
      & \cellcolor[HTML]{FFF2A8}.2727
      & \cellcolor[HTML]{C8F7D2}.3227 \\
    Mumtaz2016
      & .5000
      & .8701
      & .8972
      & \cellcolor[HTML]{FFF2A8}.8510
      & \cellcolor[HTML]{C8F7D2}.8921
      & \cellcolor[HTML]{E4D2FF}\underline{.9510} \\
    HMC
      & .2000
      & .6675
      & .7214
      & \cellcolor[HTML]{FFF2A8}.4075
      & \cellcolor[HTML]{FFF2A8}\underline{.6015}
      & \cellcolor[HTML]{FFF2A8}.4300 \\
    \bottomrule
  \end{tabular}
  \vspace{-10pt}
\end{wraptable}

After acquiring raw EEG signals from the recording device, we follow standard practice to preprocess the data with band-pass filtering, notch filtering, resampling, and amplitude normalization \cite{wang2025cbramod,zhou2025csbrain}.
The resulting dataset is denoted by $\gD=\{(\ve_i,y_i)\}_{i=1}^{N}$, where each trial $\ve_i\in\sR^{C\times T}$ is a multichannel time series and $y_i\in[K]$ is the target label.
Here, $N$ is the number of trials (samples), $C$ the number of channels, $T$ the number of time steps, and $K$ the number of classes.
The EEG decoding problem seeks a predictor $f:\sR^{C\times T}\rightarrow[K]$ that maps trials to targets, \ie, $\hat y_i=f(\ve_i)$.

In most EEG foundation model pipelines, pretraining learns an EEG encoder $\phi(\cdot)$ via self-supervised objectives, producing a representation $\phi(\ve_i)$ for each trial.
For a downstream task, one designs a task-specific head $g$ and typically fine-tunes $\phi$, yielding $\hat y_i=g(\phi(\ve_i))$ (\ie, $f=g\circ\phi$).
While this transfers general structure from pretraining, it still hinges on nontrivial per-task choices---decoder design, fine-tuning protocol, and hyperparameters---so strong performance is often not obtained ``out of the box.''
In contrast, for each target decoding problem, we distinguish a labeled support set $\gD_{\rm support}=\{(\ve_i,y_i)\}_{i=1}^{N_{\rm s}}$ from query trials $\gD_{\rm query}=\{\ve_q\}_{q=1}^{N_{\rm q}}$, whose labels are withheld at inference and used only for evaluation. We seek a support-conditioned predictor $\hat y_q=F(\ve_q\mid\gD_{\rm support})$ whose parameters remain fixed on the target dataset and whose adaptation uses neither target-validation performance nor query labels.\looseness=-1

\subsection{A Direct Frozen-ICL Attempt}
\label{sec:raw_attempt}

The most direct way to instantiate this setting is to flatten each preprocessed support trial as $\vx_i=\mathrm{vec}(\ve_i)\in\sR^{CT}$ and each query trial as $\vx_q=\mathrm{vec}(\ve_q)$, use $\{(\vx_i,y_i)\}_{i=1}^{N_{\rm s}}$ as labeled context, and query the frozen predictor with $\vx_q$. This minimal construction removes target-side gradient optimization and tests whether a general-purpose ICL predictor can exploit EEG directly before introducing any EEG-specific context operator. As \cref{tab:raw_probe} shows, raw context is promising but not consistently competitive across datasets.

Changing only the context representation shifts BAcc by $10.0$--$19.4$ percentage points in this probe, and a different representation performs best on each dataset. This sensitivity suggests that frozen ICL does not automatically isolate task-relevant EEG evidence from an arbitrary input representation: an unsuitable context may leave that evidence obscured by noise or nuisance variation, whereas a suitable context exposes temporal, spectral, or spatial structure in a form the predictor can use. Context construction is therefore the central EEG-specific adaptation problem in our framework.\looseness=-1

\subsection{A Compositional Context Interface}
\label{sec:context_interface}

\begin{figure*}[t]
  \centering
  \includegraphics[width=\linewidth]{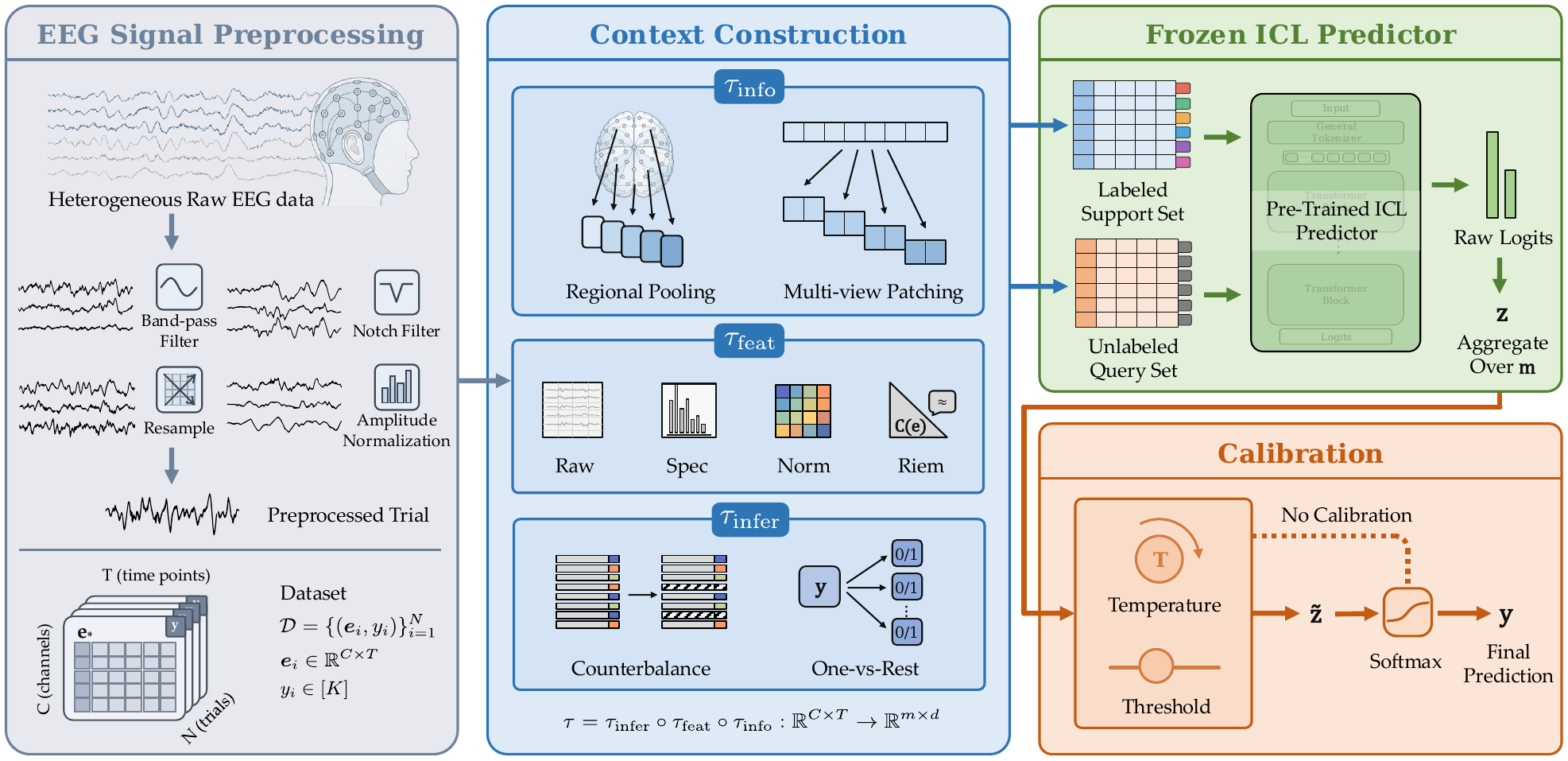}
  \vspace{-10pt}
  \caption{\textbf{Framework overview.} A context operator $\tau$ exposes EEG structure and converts each labeled support trial into one or more instances. The frozen predictor $f_{\rm icl}$ conditions on these instances to predict the query; multi-instance outputs are aggregated to the trial level, and an optional post-hoc calibration stage can adjust the operating point. No parameter of $f_{\rm icl}$ is updated for the target dataset.}
  \label{fig:framework}
  \vspace{-7pt}
\end{figure*}

As summarized in \cref{fig:framework}, our framework separates target-task adaptation into three stages. First, compositional operators construct candidate EEG contexts, and the support-only Metric route selects a construction $\tau$ for the target task without target-validation performance, query labels, or query metrics. Second, a frozen predictor $f_{\rm icl}$ conditions on the selected labeled context to produce query probabilities. Third, optional post-hoc calibration may adjust the final operating point without changing either $\tau$ or $f_{\rm icl}$.

\paragraph{Context construction and selection.} Let $\mathcal T$ denote the set of admissible compositional context operators. Using only $\gD_{\rm support}$, the Metric route selects $\tau\in\mathcal T$ for the target task; \cref{sec:contexts} details the operators and selection rule.
After fitting any required statistics exclusively on $\gD_{\rm support}$, the selected operator maps each trial to an ordered tuple $\tau(\ve_i)=(\vx_{i,1},\ldots,\vx_{i,m})\in(\sR^d)^m$. Broadcasting $y_i$ gives $\gD_{\rm support}^{\tau}=\{(\vx_{i,j},y_i)\}_{i\in[N_{\rm s}],\,j\in[m]}$, while the same operator yields $\tau(\ve_q)=(\vx_{q,1},\ldots,\vx_{q,m})$ for a query trial.

\paragraph{Frozen in-context prediction.} Let $f_{\rm icl}$ be a pretrained predictor whose parameters remain frozen. Conditioned on the selected labeled support context, it returns a class-probability vector for each query instance. We aggregate these probabilities to obtain a trial-level logit:
\begin{equation}
  \begin{aligned}
    \vz_{q}&=\mathrm{Aggregate}\big(\{f_{\rm icl}(\vx_{q,j}\mid\gD_{\rm support}^{\tau})\}_{j=1}^{m}\big)\in\sR^{K}.\!\!\!\!\!\!\!
  \end{aligned}
\end{equation}

\paragraph{Optional calibration.} 
EEG deployments can assign different costs to prediction errors: seizure detection, for example, may favor higher sensitivity (recall) to reduce missed events, even at the cost of more false alarms \cite{shoeb2009application,detti2020eeg}. We therefore expose the operating point through an optional support-only calibration stage.
Given aggregated logits $\vz_{q} \in \sR^{K}$, calibration produces adjusted logits and gives the final prediction:
\begin{equation}
  \begin{aligned}
    \tilde{\vz}_{q}=\mathrm{Calibrate}(\vz_{q}),
    \quad
    y_{q}=\argmax_{k\in[K]}\tilde{\vz}_{q,k}.
  \end{aligned}
\end{equation}

Together, these components define a tuning-free EEG foundation model in which $\tau$ provides the EEG-specific adaptation interface, $f_{\rm icl}$ supplies a reusable predictive prior, and calibration acts only after trial-level aggregation. \Cref{sec:contexts} instantiates the compositional operator space and the support-only Metric route.

\section{EEG Context Construction}
\label{sec:contexts}

\begin{figure*}[t]
  \centering
  \includegraphics[width=\linewidth]{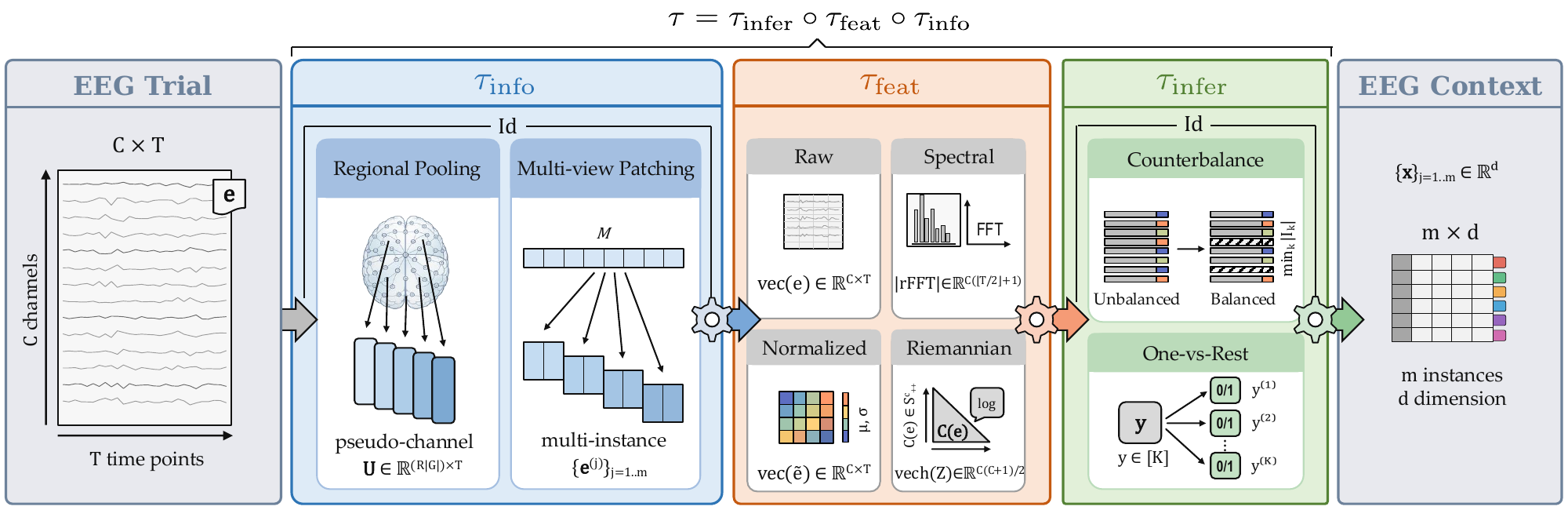}
  \vspace{-10pt}
  \caption{\textbf{EEG context construction.} The operator $\tau=\tau_{\rm infer}\circ\tau_{\rm feat}\circ\tau_{\rm info}$ combines exactly one feature type with optional temporal/spatial structure and inference-time strategies. One trial can remain one context instance or expand into local views whose predictions are aggregated. The Metric route uses four labeled-support statistics to select a context family without target-validation performance or query information.}
  \label{fig:context_construction}
  \vspace{-7pt}
\end{figure*}

Context construction is central to frozen EEG in-context prediction because it determines which task-relevant relationships are exposed between support and query trials. \Cref{fig:context_construction} summarizes the operator families and their composition.

\noindent{\bf A compositional view of EEG context construction.} We organize the EEG context interface into three operator families: (i) \emph{feature-type} operators that convert each trial or view into a vector-valued context instance, (ii) \emph{temporal and spatial information} that expose localized or anatomically organized structure, and (iii) \emph{inference-time} operators that modify support composition or label decomposition. Formally,
\begin{equation}
  \begin{aligned}
    \tau &\;\triangleq\; \tau_{\rm infer} \circ \tau_{\rm feat} \circ \tau_{\rm info}, \\
    \tau_{\rm feat}&\;\in\;\gT_{\rm feat},\;
    \tau_{\rm info}\;\in\;\gen{\gT_{\rm info}},\;
    \tau_{\rm infer}\;\in\;\gen{\gT_{\rm infer}}.
  \end{aligned}
  \label{eq:tau_composition}
\end{equation}
Exactly one feature-type operator is used, while compatible temporal/spatial and inference-time operators may be composed. The processing order is $\tau_{\rm info}\rightarrow\tau_{\rm feat}\rightarrow\tau_{\rm infer}$; for exposition, we introduce feature types first, followed by temporal/spatial structure and inference-time strategies.

\subsection{Feature Type}
\label{sec:context_feature}

Feature-type operators perform the core EEG-to-vector conversion. We define
\begin{equation}
  \gT_{\rm feat}
  =
  \{\tau_{\textsc{raw}},\tau_{\textsc{norm}},
    \tau_{\textsc{spec}},\tau_{\textsc{riem}}\}.
  \label{eq:feature_family}
\end{equation}
Each produces one context instance per input view.

\paragraph{Raw waveform context.}
The most direct operator preserves every preprocessed sample:
\begin{equation}
  \tau_{\textsc{raw}}(\ve)
  =
  \mathrm{vec}(\ve)\in\sR^{CT}.
  \label{eq:tau_raw}
\end{equation}
Time-domain neural decoders can learn useful temporal and spatial filters directly from waveforms \cite{schirrmeister2017deeplearning,lawhern2018eegnet}.
A frozen general-purpose predictor can nevertheless exploit some structure in raw waveforms through the inductive biases acquired during pretraining.

\paragraph{Per-trial normalized waveform context.}
For each EEG trial $\ve$, we jointly normalize all channel--time coordinates using statistics computed from that trial alone:
\begin{equation}
  \begin{aligned}
    \tilde e_{c,t}=\frac{e_{c,t}-\mu(\ve)}{\sqrt{\sigma^2(\ve)+\epsilon_{\rm norm}}}
    \quad
    \tau_{\textsc{norm}}(\ve)=\mathrm{vec}(\tilde\ve)\in\sR^{CT}.
  \end{aligned}
  \label{eq:tau_norm}
\end{equation}
This normalization uses neither support-set nor query-batch statistics. It suppresses trial-level offset and scale variation while preserving the relative channel--time pattern presented to the frozen predictor.

\paragraph{Spectral context.} We divide the temporal axis into $S$ non-overlapping windows of length $P$, with $T=SP$ and $\ve^{(s)}\in\sR^{C\times P}$. We apply the rFFT magnitude to each window and concatenate the resulting coefficients:
\begin{equation}
  \begin{aligned}
    \tau_{\textsc{spec}}(\ve)&=\operatorname{concat}_{s=1}^{S}\operatorname{vec}\left(|\operatorname{rFFT}(\ve^{(s)})|/P\right),\\
    \tau_{\textsc{spec}}(\ve)&\in\sR^{CS(\lfloor P/2\rfloor+1)}.
  \end{aligned}
\end{equation}
This makes oscillatory composition explicit when class evidence is more stable in frequency content than in waveform phase \cite{ang2012filterbank,yang2023biot}.

\paragraph{Covariance and Riemannian context.} We estimate an OAS covariance $\widehat{\rmC}(\ve)$ for each trial and map it to the tangent space around a Riemannian reference $\bar{\rmC}$ fitted only on support trials:
\begin{equation}
  \tau_{\textsc{riem}}(\ve)=\mathrm{vech}\big(\!\log(\bar{\rmC}^{-1/2}\widehat{\rmC}(\ve)\bar{\rmC}^{-1/2})\big)\!\in\!\sR^{C(C+1)/2}.
\end{equation}
This exposes cross-channel geometry while reducing sensitivity to exact temporal alignment \cite{barachant2012multiclass,congedo2017riemannian}.\looseness=-1

\subsection{Temporal and Spatial Structure}
\label{sec:context_structure}

Feature choice alone may still hide localized or anatomically organized evidence. We therefore define information operators which act before $\tau_{\rm feat}$:
\begin{equation}
  \gT_{\rm info} = \{\mathrm{Id},\tau_{\textsc{reg}},\tau_{\textsc{mv}}\}.
  \label{eq:information_family}
\end{equation}

\paragraph{Regional pooling.} We augment each trial with pseudo-channels obtained by fixed pooling statistics over audited anatomical channel groups, and then apply the Riemannian feature operator. This deterministic augmentation exposes coarse regional organization without learning additional weights \cite{zhou2025csbrain,fang2025neuript}.

\paragraph{Local temporal views.} We split each trial into $m$ temporal windows and treat each window as a context instance. Support labels are repeated across views, while query probabilities are averaged back to the trial level. This exposes localized evidence without learning a temporal aggregation module \cite{wang2024eegpt,zhou2025csbrain}.

\subsection{Inference-Time Strategies}
\label{sec:context_inference}

Inference-time strategies modify the labeled context or class decomposition without changing the frozen predictor. We define
\begin{equation}
  \gT_{\rm infer}=\{\mathrm{Id},\tau_{\textsc{cb}},\tau_{\textsc{ovr}}\}.
\end{equation}

\paragraph{Counterbalanced support.} We use seeded subsampling to retain the same number of support trials from every class. Context budgets are then enforced over complete trials after accounting for any trial-to-row expansion introduced by the selected context operator.

\paragraph{One-versus-rest prediction.} For a $K$-class task, we construct $K$ binary labeled contexts and obtain one positive-class probability from each frozen contextual prediction. Normalizing these scores yields the multiclass distribution. This decomposition can simplify a large native label space, but requires $K$ prediction passes.

\begin{figure}[t]
  \centering
  \includegraphics[width=\linewidth]{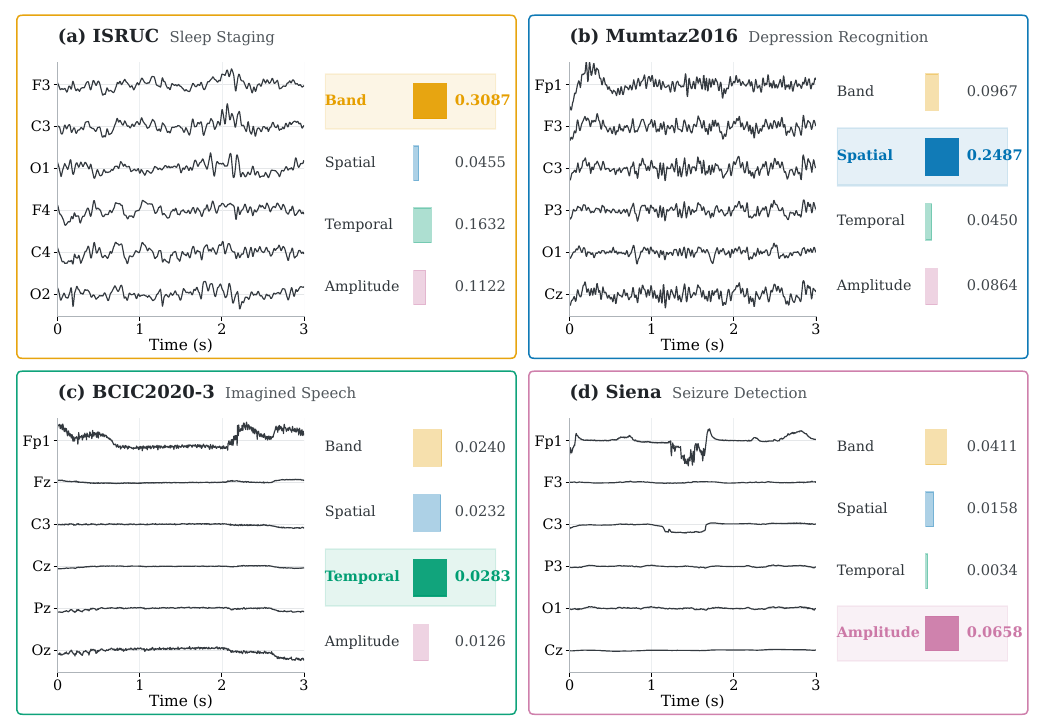}
  \vspace{-10pt}
  \caption{\textbf{Support-only Metric routing.} Each panel shows an EEG trial and four class-separation scores from the labeled, class-balanced support subset. Routing uses no validation or query information.}
  \label{fig:routing_metric_examples}
  \vspace{-7pt}
\end{figure}

\subsection{Support-Only Context Selection}
\label{sec:metric_route}
\paragraph{Support-derived separation scores.}
To select $\tau$ from labeled support without evaluating candidate contexts, the Metric route summarizes every support trial using four low-dimensional descriptor groups. The \emph{band} group contains relative $\delta$-, $\theta$-, $\alpha$-, $\beta$-, and $\gamma$-band power \cite{ang2012filterbank,saeidi2021neuraldecoding}; the \emph{spatial} group contains mean absolute off-diagonal channel correlation and normalized correlation effective rank \cite{roy2007effective}; the \emph{temporal} group contains lag-one autocorrelation, normalized line length, and nonstationarity \cite{esteller2001line}; and the \emph{amplitude} group contains RMS, kurtosis, and crest factor \cite{saeidi2021neuraldecoding}.
Let $\mathcal G=\{\mathrm{band},\mathrm{spatial},\mathrm{temporal},\mathrm{amplitude}\}$. For each $g\in\mathcal G$, $\vv_i^{(g)}\in\sR^{d_g}$ denotes the descriptor of support trial $i$, and $\widetilde\vv_i^{(g)}$ denotes its robustly standardized version. We present the non-degenerate case with $n$ support trials from $K$ represented classes. Let $n_k$ be the number of trials in class $k$, $\vmu_k^{(g)}$ its mean standardized descriptor, and $\vmu^{(g)}$ the pooled mean.
We first measure the between-class dispersion of descriptor group $g$:
\begin{equation}
  B_g=\frac{1}{d_g}\sum_{k=1}^{K}n_k\|\vmu_k^{(g)}-\vmu^{(g)}\|_2^2.
  \label{eq:support_between_dispersion}
\end{equation}
We separately measure its within-class dispersion:
\begin{equation}
  W_g=\frac{1}{d_g}\sum_{k=1}^{K}\sum_{i:y_i=k}\|\widetilde\vv_i^{(g)}-\vmu_k^{(g)}\|_2^2.
  \label{eq:support_within_dispersion}
\end{equation}
Their degrees-of-freedom-normalized ratio gives the raw class-separation statistic
\begin{equation}
  r_g=\frac{B_g/n}{\max\!\left\{W_g/(n-K),\epsilon\right\}},
  \label{eq:support_separation_ratio}
\end{equation}
where $\epsilon=10^{-8}$ prevents division by zero. We then log-compress this statistic to obtain the routing score
\begin{equation}
  s_g=\log\!\left(1+\min\{r_g,r_{\max}\}\right),\qquad r_{\max}=10^6.
  \label{eq:support_separation}
\end{equation}
A larger $s_g$ indicates that descriptor group $g$ exhibits stronger between-class separation relative to its within-class variation.

\paragraph{Deterministic context mapping.}
We found counterbalanced support consistently beneficial across the evaluated contexts, so $\tau_{\textsc{cb}}$ is enabled for every routed context. Let $g^\star=\argmax_g s_g$ denote the dominant support descriptor. The three operator slots are selected as:
\begin{equation}
  \begin{aligned}
  & \tau_{\rm feat}(g^\star) =
    \begin{cases}
      \tau_{\textsc{norm}}, & g^\star=\mathrm{band}\ \text{and}\ s_{\mathrm{temporal}} \leq s_{\mathrm{amplitude}}, \\
      \tau_{\textsc{spec}}, & g^\star=\mathrm{band}\ \text{and}\ s_{\mathrm{temporal}} > s_{\mathrm{amplitude}}, \\
      \tau_{\textsc{riem}}, & g^\star\in\{\mathrm{spatial},\mathrm{amplitude}\}, \\
      \tau_{\textsc{raw}},  & g^\star=\mathrm{temporal},
    \end{cases} \\
  & \tau_{\rm info}(g^\star) =
    \begin{cases}
      \tau_{\textsc{reg}},  & g^\star=\mathrm{amplitude}, \\
      \tau_{\textsc{mv}},   & g^\star=\mathrm{temporal}, \\
      \mathrm{Id},          & \text{otherwise},
    \end{cases} \\
  & \tau_{\rm infer}(g^\star,K) =
    \begin{cases}
      \tau_{\textsc{ovr}}\circ\tau_{\textsc{cb}}, & K>2\ \text{and}\ g^\star\neq\mathrm{band}, \\
      \tau_{\textsc{cb}},   & \text{otherwise}.
    \end{cases}
  \end{aligned}
  \label{eq:metric_operator_selection}
\end{equation}
Exact ties follow the fixed order band, spatial, temporal, and amplitude. If every score is undefined, we bypass the mapping and use $\tau_{\rm feat}=\tau_{\textsc{raw}}$, $\tau_{\rm info}=\mathrm{Id}$, and $\tau_{\rm infer}=\tau_{\textsc{cb}}$. \Cref{fig:routing_metric_examples} visualizes representative selections.
The resulting context choices and their remaining gap to the Oracle are evaluated in \cref{sec:experiment}.

\begin{figure*}[t]
  \centering
  \begin{minipage}[t]{.495\textwidth}
    \centering
    \includegraphics[width=\linewidth]{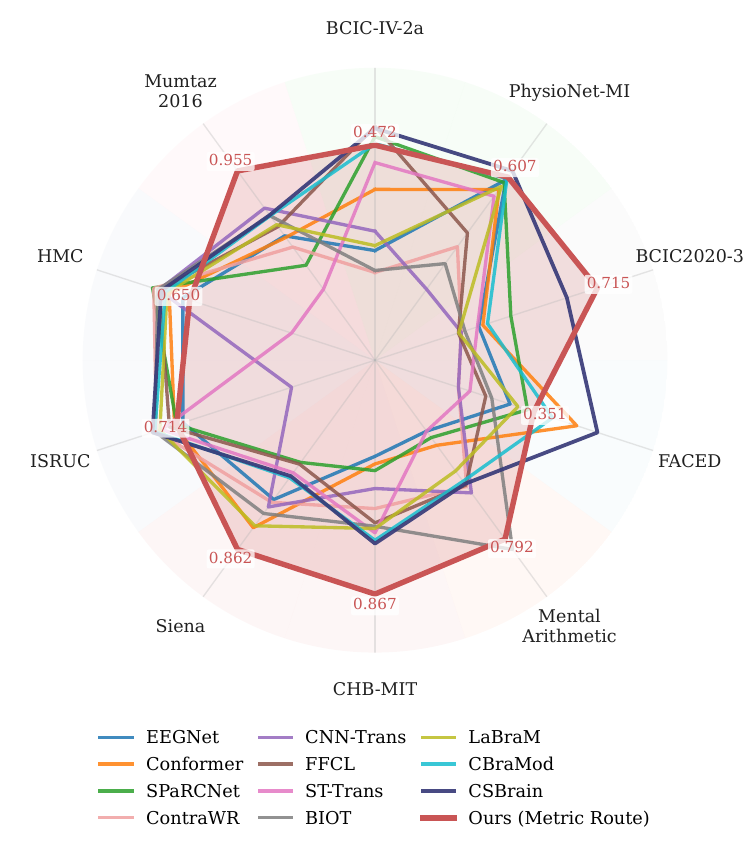}
  \end{minipage}\hfill
  \begin{minipage}[t]{.495\textwidth}
    \centering
    \includegraphics[width=\linewidth]{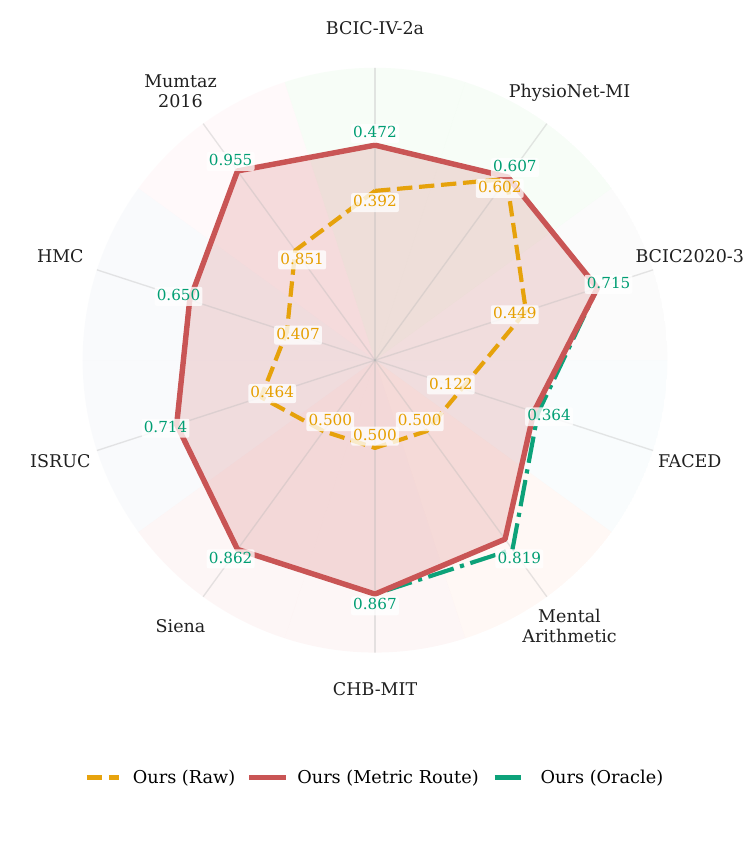}
  \end{minipage}
  \vspace{-5pt}
  \caption{\textbf{Overall performance and context adaptation across ten datasets.} \textbf{Left:} Ours (Metric) compared with eleven locally reproduced EEG baselines, with its numerical BAcc annotated. \textbf{Right:} Raw, Metric, and inspection-Oracle contexts, with numerical BAcc annotated for Raw and Oracle. Metric routing consistently improves over Raw and closely approaches the inspection Oracle. Radial positions use per-dataset min--max scaling shared across both panels.}
  \label{fig:result_radar}
  \vspace{-7pt}
\end{figure*}

\section{Experiments and Analysis}
\label{sec:experiment}

In this section, we evaluate overall decoding performance, support-only context adaptation, and threshold calibration. Additional results on predictor variants, context budgets, per-dataset metrics, and runtime are provided in the appendix.

\paragraph{Experimental Setup.} We evaluate ten established EEG datasets spanning seven paradigms: motor imagery, imagined speech, emotion recognition, mental stress, seizure detection, sleep staging, and mental-disorder diagnosis \cite{brunner2008bci,schalk2004bci2000,jeong20222020,chen2023large,zyma2019electroencephalograms,shoeb2009application,detti2020eeg,khalighi2016isrucsleep,alvarez2021inter,mumtaz2016mdd}. The training split provides labeled support and the held-out test split provides query trials. We do not use validation split in our ICL method. We report balanced accuracy (BAcc) as the primary metric; context construction and routing use neither target-validation performance nor query information.

We instantiate the frozen predictor with TabPFN-2.5 \cite{grinsztajn2025tabpfn25}. We compare against seven specialized decoders \cite{lawhern2018eegnet,song2021transformer,yang2023contrawr,peh2022cnntransformer,li2022ffcl,jing2023sparcnet,song2023eegconformer} and four pretrained EEG models \cite{yang2023biot,jiang2024labram,wang2025cbramod,zhou2025csbrain}.
Full details on datasets, preprocessing, baselines, context budgets, additional metrics, predictor variants, and runtime are provided in the appendix. \looseness=-1

\Cref{tab:dataset_contexts} shows the configurations selected independently from each dataset's labeled support set. Both sleep-staging datasets select spectral features, whereas both motor-imagery datasets select normalized waveforms.

\begin{wraptable}{R}{.5\textwidth}
\captionsetup{hypcap=false}
\begin{minipage}{\linewidth}
  \vspace{-13.4pt}
  \centering
  \caption{\textbf{EEG context configurations by paradigm.} For each paradigm represented by two datasets, the support-only Metric route selected the same three operators for both. $\mathrm{Id}$ denotes the identity operator.}
  \label{tab:dataset_contexts}
  \small
  \setlength{\tabcolsep}{3pt}
  \begin{tabular}{@{}llccc@{}}
    \toprule
    \textbf{Paradigm} & \textbf{Dataset} & $\tau_{\rm feat}$ & $\tau_{\rm info}$ & $\tau_{\rm infer}$ \\
    \midrule
    \multirow{2}{*}{Motor imagery} & BCIC-IV-2a & \multirow{2}{*}{$\tau_{\textsc{norm}}$} & \multirow{2}{*}{$\mathrm{Id}$} & \multirow{2}{*}{$\tau_{\textsc{cb}}$} \\
    & PhysioNet-MI & & & \\
    \midrule
    Imagined speech & BCIC2020-3 & $\tau_{\textsc{raw}}$ & $\tau_{\textsc{mv}}$ & $\tau_{\textsc{ovr}}\circ\tau_{\textsc{cb}}$ \\
    \midrule
    \multirow{2}{*}{Seizure detection} & CHB-MIT & \multirow{2}{*}{$\tau_{\textsc{riem}}$} & \multirow{2}{*}{$\tau_{\textsc{reg}}$} & \multirow{2}{*}{$\tau_{\textsc{cb}}$} \\
    & Siena & & & \\
    \midrule
    \makecell[l]{Emotion\\recognition} & FACED & $\tau_{\textsc{raw}}$ & $\tau_{\textsc{mv}}$ & $\tau_{\textsc{ovr}}\circ\tau_{\textsc{cb}}$ \\
    \midrule
    \multirow{2}{*}{Sleep staging} & HMC & \multirow{2}{*}{$\tau_{\textsc{spec}}$} & \multirow{2}{*}{$\mathrm{Id}$} & \multirow{2}{*}{$\tau_{\textsc{cb}}$} \\
    & ISRUC & & & \\
    \midrule
    Mental stress & MentalArithmetic & $\tau_{\textsc{riem}}$ & $\tau_{\textsc{reg}}$ & $\tau_{\textsc{cb}}$ \\
    \midrule
    Mental disorder & Mumtaz2016 & $\tau_{\textsc{riem}}$ & $\mathrm{Id}$ & $\tau_{\textsc{cb}}$ \\
    \bottomrule
  \end{tabular}
\end{minipage}\par\vspace{16pt}
\begin{minipage}{\linewidth}
  \centering
  \small
  \caption{\textbf{Main comparison across ten EEG datasets.} The best and second-best results are shown in \textbf{bold} and \underline{underlined}. Average ranks are computed independently on each dataset using BAcc. Without target-task fine-tuning or validation-based model selection, Ours (Metric) achieves the highest mean BAcc and the second-best average rank.}
  \label{tab:main_results}
  \setlength{\tabcolsep}{3pt}
  \begin{tabular}{clcc}
    \toprule
    \textbf{Setting} & \textbf{Method} & \textbf{Avg. B-Acc.}$\uparrow$ & \textbf{Avg. Rank}$\downarrow$ \\
    \midrule
    \multirow{7}{*}{Specialized}
      & EEGNet          & .5400 & 8.65 \\
      & ST-Transformer  & .5202 & 8.65 \\
      & ContraWR        & .5353 & 7.70 \\
      & CNN-Transformer & .4915 & 7.60 \\
      & FFCL            & .5617 & 7.30 \\
      & SPaRCNet        & .5760 & 7.00 \\
      & EEGConformer    & .5905 & 6.90 \\
    \midrule
    \multirow{4}{*}{\makecell[c]{Pre-training\\\&\\Fine-tuning}}
      & LaBraM          & .5831 & 6.70 \\
      & BIOT            & .5709 & 5.95 \\
      & CBraMod         & .6191 & 4.55 \\
      & CSBrain         & \underline{.6707} & \textbf{3.30} \\
    \midrule
    \textbf{Tuning-free} & \textbf{Ours (Metric)} & \textbf{.6985} & \underline{3.70} \\
    \bottomrule
  \end{tabular}
\end{minipage}\par\vspace{16pt}
\begin{minipage}{\linewidth}
  \centering
  \small
  \caption{\textbf{Support-only threshold calibration on Siena with TabPFN-3.} Thresholds are selected from participant-grouped support OOF predictions and frozen before test evaluation.}
  \label{tab:main_siena_calibration}
  \setlength{\tabcolsep}{2.6pt}
  \begin{tabular}{ccccc}
    \toprule
    \textbf{Operating Point} & \textbf{Threshold} & \textbf{B-Acc.}$\uparrow$ & \textbf{Recall}$\uparrow$ & \textbf{Specificity}$\uparrow$ \\
    \midrule
    Default         & .5000 & .8534 & .7487 & \textbf{.9581} \\
    OOF $\geq .60$ & .3698 & \textbf{.8807} & .8717 & .8898 \\
    OOF $\geq .95$ & .0621 & .6881 & \textbf{.9572} & .4189 \\
    \bottomrule
  \end{tabular}
\end{minipage}
\vspace{-70pt}
\end{wraptable}

\paragraph{Overall Comparison.} \Cref{tab:main_results} shows that Ours (Metric) achieves the highest mean BAcc of .6985, exceeding the strongest baseline, CSBrain, by 2.78 percentage points, while attaining the second-best average rank of 3.70. Together with the broad per-dataset profile in the left panel of \cref{fig:result_radar}, these results demonstrate that a frozen general-purpose predictor can match or exceed specialized and fine-tuned EEG models without target-task gradient updates or validation-based model selection.

\paragraph{Effectiveness of Support-Only Context Adaptation.} The right panel of \cref{fig:result_radar} shows that Metric routing improves mean BAcc from .4788 with Raw context to .6985 and improves every evaluated dataset, demonstrating that context adaptation is essential to the effectiveness of the frozen predictor. Ours (Metric) further matches the current inspection Oracle on eight of ten datasets, with an average gap of only .0041, showing that support-derived statistics can identify highly effective contexts without target-validation selection. \looseness=-1

\paragraph{Threshold Calibration.} In seizure detection, missed seizure events can be more consequential than false alarms, making high recall an important operating requirement. As shown in \cref{tab:main_siena_calibration}, we use Siena as a case study to examine whether labeled support data can also adapt the decision threshold after context selection. Selecting a threshold that attains $.60$ recall on participant-grouped support OOF predictions raises test recall from .7487 to .8717 and BAcc from .8534 to .8807, without using target-validation data or query labels. A more aggressive OOF recall target of $.95$ further increases test recall to .9572, while reducing specificity to .4189, demonstrating a controllable sensitivity--specificity trade-off. Because calibration changes only the threshold applied to fixed prediction scores, AUC remains .9237.

\section{Conclusion}
\label{sec:conclusion}

We introduce a general framework for constructing tuning-free EEG foundation models by coupling EEG-specific context construction with a frozen general-purpose ICL predictor, making labeled context the interface for target-task adaptation. 
Having identified context construction as the central challenge, we develop a family of complementary contexts and a support-only metric-based adaptation rule that selects an appropriate context for each target task without validation performance or query information. 
Across ten datasets spanning seven EEG decoding paradigms, our Metric route achieves the highest mean BAcc among eleven specialized and fine-tuned EEG baselines, improves over raw context on every dataset, and closely approaches the inspection Oracle. These results demonstrate that EEG-specific context construction and support-derived adaptation enable a frozen general-purpose ICL predictor to serve as an effective EEG decoder without task-specific parameter updates or validation-based model selection. 
More broadly, our framework provides a practical path toward EEG foundation models in which labeled context, rather than downstream optimization, specifies the target decoding task.\looseness=-1

\bibliography{ref}

@article{brunner2008bci,
  author       = {Brunner, C and
                  Leeb, R and
                  M{\"u}ller-Putz, GR and
                  Schl{\"o}gl, A and
                  Pfurtscheller, G},
  title        = {BCI Competition 2008--Graz data set A},
  journal      = {Institute for knowledge discovery (laboratory of brain-computer interfaces)},
  volume       = {16},
  pages        = {1-6},
  year         = {2008},
}

@article{schalk2004bci2000,
  author       = {Gerwin Schalk and
                  Dennis J. McFarland and
                  Thilo Hinterberger and
                  Niels Birbaumer and
                  Jonathan R. Wolpaw},
  title        = {{BCI2000:} a general-purpose brain-computer interface {(BCI)} system},
  journal      = {{IEEE} Trans. Biomed. Eng.},
  volume       = {51},
  number       = {6},
  pages        = {1034--1043},
  year         = {2004}
}

@article{chen2023large,
  author       = {Chen, Jingjing and
                  Wang, Xiaobin and
                  Huang, Chen and
                  Hu, Xin and
                  Shen, Xinke and
                  Zhang, Dan},
  title        = {A large finer-grained affective computing EEG dataset},
  journal      = {Sci. Data},
  volume       = {10},
  number       = {1},
  pages        = {740},
  year         = {2023},
}

@phdthesis{shoeb2009application,
  author       = {Shoeb, Ali Hossam},
  title        = {Application of machine learning to epileptic seizure onset detection and treatment},
  year         = {2009},
  school       = {Massachusetts Institute of Technology},
}

@article{detti2020eeg,
  author       = {Detti, Paolo and
                  Vatti, Giampaolo and
                  Zabalo Manrique de Lara, Garazi},
  title        = {EEG synchronization analysis for seizure prediction: A study on data of noninvasive recordings},
  journal      = {Processes},
  volume       = {8},
  number       = {7},
  pages        = {846},
  year         = {2020},
}

@article{khalighi2016isrucsleep,
  author       = {Sirvan Khalighi and
                  Teresa Sousa and
                  Jos{\'{e}} Moutinho dos Santos and
                  Urbano Nunes},
  title        = {ISRUC-Sleep: {A} comprehensive public dataset for sleep researchers},
  journal      = {Comput. Methods Programs Biomed.},
  volume       = {124},
  pages        = {180--192},
  year         = {2016}
}

@article{alvarez2021inter,
  author       = {Alvarez-Estevez, Diego and
                  Rijsman, Roselyne M},
  title        = {Inter-database validation of a deep learning approach for automatic sleep scoring},
  journal      = {{PLOS} ONE},
  volume       = {16},
  number       = {8},
  pages        = {e0256111},
  year         = {2021},
}

@article{jeong20222020,
  author       = {Jeong, Ji-Hoon and
                  Cho, Jeong-Hyun and
                  Lee, Young-Eun and
                  Lee, Seo-Hyun and
                  Shin, Gi-Hwan and
                  Kweon, Young-Seok and
                  Mill{\'a}n, Jos{\'e} del R and
                  M{\"u}ller, Klaus-Robert and
                  Lee, Seong-Whan},
  title        = {2020 International brain--computer interface competition: A review},
  journal       = {Front. Hum. Neurosci.},
  volume        = {16},
  pages         = {898300},
  year          = {2022},
}

@article{zyma2019electroencephalograms,
  author       = {Igor Zyma and
                  Sergii Tukaiev and
                  Ivan Seleznov and
                  Ken Kiyono and
                  Anton Popov and
                  Mariia Chernykh and
                  Oleksii Shpenkov},
  title        = {Electroencephalograms during Mental Arithmetic Task Performance},
  journal      = {Data},
  volume       = {4},
  number       = {1},
  pages        = {14},
  year         = {2019}
}

@article{mumtaz2016mdd,
  author       = {Mumtaz, Wajid},
  title        = {MDD patients and healthy controls EEG data (new)},
  journal      = {figshare, Dataset},
  year         = {2016}
}

@article{lotte2018review,
  author       = {Lotte, F and
                  Bougrain, L and
                  Cichocki, A and
                  Clerc, M and
                  Congedo, M and
                  Rakotomamonjy, A and
                  Yger, F},
  title        = {A review of classification algorithms for EEG-based brain–computer interfaces: a 10 year update},
  journal      = {J. Neural Eng.},
  volume       = {15},
  number       = {3},
  pages        = {031005},
  year         = {2018},
}

@article{congedo2017riemannian,
  author       = {Marco Congedo and
                  Alexandre Barachant and
                  Rajendra Bhatia},
  title        = {Riemannian geometry for EEG-based brain-computer interfaces; a primer and a review},
  journal      = {Brain Comput. Interfaces},
  volume       = {4},
  number       = {3},
  pages        = {155--174},
  year         = {2017},
}

@article{barachant2012multiclass,
  author       = {Barachant, Alexandre and
                  Bonnet, Stéphane and
                  Congedo, Marco and
                  Jutten, Christian},
  journal      = {{IEEE} Trans. Biomed. Eng.}, 
  title        = {Multiclass Brain–Computer Interface Classification by Riemannian Geometry}, 
  year         = {2012},
  volume       = {59},
  number       = {4},
  pages        = {920-928},
}

@article{chevallier2024moabb,
  author       = {Sylvain Chevallier and
                  Igor Carrara and
                  Bruno Aristimunha and
                  Pierre Guetschel and
                  Sara Sedlar and
                  Bruna Lopes and
                  S{\'{e}}bastien Velut and
                  Salim Khazem and
                  Thomas Moreau},
  title        = {The largest EEG-based {BCI} reproducibility study for open science:
                  the {MOABB} benchmark},
  journal      = {CoRR},
  volume       = {abs/2404.15319},
  year         = {2024}
}

@article{schirrmeister2017deeplearning,
  author = {Schirrmeister, Robin Tibor and Springenberg, Jost Tobias and Fiederer, Lukas Dominique Josef and Glasstetter, Martin and Eggensperger, Katharina and Tangermann, Michael and Hutter, Frank and Burgard, Wolfram and Ball, Tonio},
  title = {Deep learning with convolutional neural networks for EEG decoding and visualization},
  journal = {Hum. Brain Mapp.},
  volume = {38},
  number = {11},
  pages = {5391-5420},
  year = {2017}
}

@article{lawhern2018eegnet,
  author       = {Lawhern, Vernon J and
                  Solon, Amelia J and
                  Waytowich, Nicholas R and
                  Gordon, Stephen M and
                  Hung, Chou P and
                  Lance, Brent J},
  title        = {EEGNet: a compact convolutional neural network for EEG-based brain–computer interfaces},
  journal      = {J. Neural Eng.},
  volume       = {15},
  number       = {5},
  pages        = {056013},
  year         = {2018},
}

@article{song2023eegconformer,
  author       = {Song, Yonghao and
                  Zheng, Qingqing and
                  Liu, Bingchuan and
                  Gao, Xiaorong},
  title        = {EEG Conformer: Convolutional Transformer for EEG Decoding and Visualization}, 
  journal      = {{IEEE} Trans. Neural Syst. Rehabilitation Eng.},
  volume       = {31},
  pages        = {710-719},
  year         = {2023},
}

@article{jing2023sparcnet,
  author       = {Jin Jing and
                  Wendong Ge and
                  Shenda Hong and
                  Marta Bento Fernandes and
                  Zhen Lin and
                  Chaoqi Yang and
                  Sungtae An and
                  Aaron F. Struck and
                  Aline Herlopian and
                  Ioannis Karakis and
                  Jonathan J. Halford and
                  Marcus C. Ng and
                  Emily L. Johnson and
                  Brian L. Appavu and
                  Rani A. Sarkis and
                  Gamaleldin Osman and
                  Peter W. Kaplan and
                  Monica B. Dhakar and
                  Lakshman Arcot Jayagopal and
                  Zubeda Sheikh and
                  Olga Taraschenko and
                  Sarah Schmitt and
                  Hiba A. Haider and
                  Jennifer A. Kim and
                  Christa B. Swisher and
                  Nicolas Gaspard and
                  Mackenzie C. Cervenka and
                  Andres A. Rodriguez Ruiz and
                  Jong Woo Lee and
                  Mohammad Tabaeizadeh and
                  Emily J. Gilmore and
                  Kristy Nordstrom and
                  Ji Yeoun Yoo and
                  Manisha G. Holmes and
                  Susan T. Herman and
                  Jennifer A. Williams and
                  Jay Pathmanathan and
                  Fábio A. Nascimento and
                  Ziwei Fan and
                  Samaneh Nasiri and
                  Mouhsin M. Shafi and
                  Sydney S. Cash and
                  Daniel B. Hoch and
                  Andrew J. Cole and
                  Eric S. Rosenthal and
                  Sahar F. Zafar and
                  Jimeng Sun and
                  M. Brandon Westover},
  title        = {Development of Expert-Level Classification of Seizures and Rhythmic and Periodic Patterns During EEG Interpretation},
  journal      = {Neurology},
  volume       = {100},
  number       = {17},
  pages        = {e1750-e1762},
  year         = {2023},
}

@Article{yang2023contrawr,
  author       = {Yang, Chaoqi and
                  Xiao, Cao and
                  Westover, M Brandon and
                  Sun, Jimeng},
  title        = {Self-Supervised Electroencephalogram Representation Learning for Automatic Sleep Staging: Model Development and Evaluation Study},
  journal      = {JMIR AI},
  volume       = {2},
  pages        = {e46769},
  year         = {2023},
}

@inproceedings{peh2022cnntransformer,
  author       = {Wei Yan Peh and
                  Yuanyuan Yao and
                  Justin Dauwels},
  title        = {Transformer Convolutional Neural Networks for Automated Artifact Detection
                  in Scalp {EEG}},
  booktitle    = {{EMBC}},
  pages        = {3599--3602},
  year         = {2022}
}

@article{li2022ffcl,
  author       = {Hongli Li and
                  Man Ding and
                  Ronghua Zhang and
                  Chunbo Xiu},
  title        = {Motor imagery {EEG} classification algorithm based on {CNN-LSTM} feature
                  fusion network},
  journal      = {Biomed. Signal Process. Control},
  volume       = {72},
  number       = {Part},
  pages        = {103342},
  year         = {2022}
}

@article{song2021transformer,
  author       = {Yonghao Song and
                  Xueyu Jia and
                  Lie Yang and
                  Longhan Xie},
  title        = {Transformer-based Spatial-Temporal Feature Learning for {EEG} Decoding},
  journal      = {CoRR},
  volume       = {abs/2106.11170},
  year         = {2021}
}

@inproceedings{yang2023biot,
  author       = {Chaoqi Yang and
                  M. Brandon Westover and
                  Jimeng Sun},
  title        = {{BIOT:} Biosignal Transformer for Cross-data Learning in the Wild},
  booktitle    = {NeurIPS},
  pages        = {78240--78260},
  year         = {2023}
}

@inproceedings{jiang2024labram,
  author       = {Wei{-}Bang Jiang and
                  Li{-}Ming Zhao and
                  Bao{-}Liang Lu},
  title        = {Large Brain Model for Learning Generic Representations with Tremendous
                  {EEG} Data in {BCI}},
  booktitle    = {{ICLR}},
  year         = {2024}
}

@inproceedings{wang2025cbramod,
  author       = {Jiquan Wang and
                  Sha Zhao and
                  Zhiling Luo and
                  Yangxuan Zhou and
                  Haiteng Jiang and
                  Shijian Li and
                  Tao Li and
                  Gang Pan},
  title        = {CBraMod: {A} Criss-Cross Brain Foundation Model for {EEG} Decoding},
  booktitle    = {{ICLR}},
  year         = {2025}
}

@article{zhou2025csbrain,
  author       = {Yuchen Zhou and
                  Jiamin Wu and
                  Zichen Ren and
                  Zhouheng Yao and
                  Weiheng Lu and
                  Kunyu Peng and
                  Qihao Zheng and
                  Chunfeng Song and
                  Wanli Ouyang and
                  Chao Gou},
  title        = {CSBrain: {A} Cross-scale Spatiotemporal Brain Foundation Model for
                  {EEG} Decoding},
  journal      = {CoRR},
  volume       = {abs/2506.23075},
  year         = {2025}
}

@article{xiao2025brainomni,
  author       = {Qinfan Xiao and
                  Ziyun Cui and
                  Chi Zhang and
                  Siqi Chen and
                  Wen Wu and
                  Andrew Thwaites and
                  Alexandra Woolgar and
                  Bowen Zhou and
                  Chao Zhang},
  title        = {BrainOmni: {A} Brain Foundation Model for Unified {EEG} and {MEG}
                  Signals},
  journal      = {CoRR},
  volume       = {abs/2505.18185},
  year         = {2025}
}

@inproceedings{hollmann2022tabpfn,
  author       = {Noah Hollmann and
                  Samuel M{\"{u}}ller and
                  Katharina Eggensperger and
                  Frank Hutter},
  title        = {TabPFN: {A} Transformer That Solves Small Tabular Classification Problems
                  in a Second},
  booktitle    = {{ICLR}},
  year         = {2023}
}

@article{hollmann2025accurate,
  author       = {Hollmann, Noah and
                  Müller, Samuel and
                  Purucker, Lennart and
                  Krishnakumar, Arjun and
                  Körfer, Max and
                  Hoo, Shi Bin and
                  Schirrmeister, Robin Tibor and
                  Hutter, Frank},
  title        = {Accurate predictions on small data with a tabular foundation model},
  journal      = {Nature},
  volume       = {637},
  pages        = {319--326},
  year         = {2025}
}

@inproceedings{qu2025tabicl,
  author       = {Jingang Qu and
                  David Holzm{\"{u}}ller and
                  Ga{\"{e}}l Varoquaux and
                  Marine Le Morvan},
  title        = {TabICL: {A} Tabular Foundation Model for In-Context Learning on Large
                  Data},
  booktitle    = {{ICML}},
  pages        = {50817--50847},
  year         = {2025}
}

@article{grinsztajn2025tabpfn25,
  author       = {L{\'{e}}o Grinsztajn and
                  Klemens Fl{\"{o}}ge and
                  Oscar Key and
                  Felix Birkel and
                  Philipp Jund and
                  Brendan Roof and
                  Benjamin J{\"{a}}ger and
                  Dominik Safaric and
                  Simone Alessi and
                  Adrian Hayler and
                  Mihir Manium and
                  Rosen Yu and
                  Felix Jablonski and
                  Shi Bin Hoo and
                  Anurag Garg and
                  Jake Robertson and
                  Magnus B{\"{u}}hler and
                  Vladyslav Moroshan and
                  Lennart Purucker and
                  Clara Cornu and
                  Lilly Charlotte Wehrhahn and
                  Alessandro Bonetto and
                  Bernhard Sch{\"{o}}lkopf and
                  Sauraj Gambhir and
                  Noah Hollmann and
                  Frank Hutter},
  title        = {TabPFN-2.5: Advancing the State of the Art in Tabular Foundation Models},
  journal      = {CoRR},
  volume       = {abs/2511.08667},
  year         = {2025}
}

@article{saha2020variability,
  author       = {Simanto Saha and Mathias Baumert},
  title        = {Intra- and Inter-subject Variability in {EEG}-Based Sensorimotor Brain Computer Interface: A Review},
  journal      = {Front. Comput. Neurosci.},
  volume       = {13},
  pages        = {87},
  year         = {2020}
}

@article{wu2022transfer,
  author       = {Dongrui Wu and Yifan Xu and Bao{-}Liang Lu},
  title        = {Transfer Learning for {EEG}-Based Brain--Computer Interfaces: A Review of Progress Made Since 2016},
  journal      = {{IEEE} Trans. Cogn. Dev. Syst.},
  volume       = {14},
  number       = {1},
  pages        = {4--19},
  year         = {2022}
}

@article{huang2021calibration,
  author       = {Xueqian Huang and Yilu Xu and Junhua Hua and Wanzeng Yi and Huibin Yin and Rong Hu and Shuyue Wang},
  title        = {A Review on Signal Processing Approaches to Reduce Calibration Time in {EEG}-Based Brain--Computer Interface},
  journal      = {Front. Neurosci.},
  volume       = {15},
  pages        = {733546},
  year         = {2021}
}

@inproceedings{bomatter2025diversity,
  author       = {Philipp Bomatter and Henry Gouk},
  title        = {Is Limited Participant Diversity Impeding {EEG}-Based Machine Learning?},
  booktitle    = {NeurIPS},
  pages        = {117447--117474},
  year         = {2025}
}

@article{ang2012filterbank,
  author       = {Kai Keng Ang and Zheng Yang Chin and Chuanchu Wang and Cuntai Guan and Haihong Zhang},
  title        = {Filter Bank Common Spatial Pattern Algorithm on {BCI} Competition {IV} Datasets 2a and 2b},
  journal      = {Front. Neurosci.},
  volume       = {6},
  pages        = {39},
  year         = {2012}
}

@inproceedings{wang2024eegpt,
  author       = {Guagnyu Wang and Wenchao Liu and Yuhong He and Cong Xu and Lin Ma and Haifeng Li},
  title        = {{EEGPT}: Pretrained Transformer for Universal and Reliable Representation of {EEG} Signals},
  booktitle    = {NeurIPS},
  pages        = {39249--39280},
  year         = {2024}
}

@inproceedings{elouahidi2025reve,
  author       = {Yassine {El Ouahidi} and Jonathan Lys and Philipp Th{\"o}lke and Nicolas Farrugia and Bastien Pasdeloup and Vincent Gripon and Karim Jerbi and Giulia Lioi},
  title        = {{REVE}: A Foundation Model for {EEG}---Adapting to Any Setup with Large-Scale Pretraining on 25,000 Subjects},
  booktitle    = {NeurIPS},
  pages        = {22541--22577},
  year         = {2025}
}

@inproceedings{doner2025luna,
  author       = {Berkay D{\"o}ner and Thorir Mar Ingolfsson and Luca Benini and Yawei Li},
  title        = {{LUNA}: Efficient and Topology-Agnostic Foundation Model for {EEG} Signal Analysis},
  booktitle    = {NeurIPS},
  pages        = {70682--70708},
  year         = {2025}
}

@inproceedings{fang2025neuript,
  author       = {Zitao Fang and Chenxuan Li and Hongting Zhou and Shuyang Yu and Guodong Du and Ashwaq Qasem and Yang Lu and Jing Li and Junsong Zhang and Sim Kuan Goh},
  title        = {{NeurIPT}: Foundation Model for Neural Interfaces},
  booktitle    = {NeurIPS},
  pages        = {150819--150861},
  year         = {2025}
}

@inproceedings{lee2025capable,
  author       = {Na Lee and Konstantinos Barmpas and Yannis Panagakis and Dimitrios Adamos and Nikolaos Laskaris and Stefanos Zafeiriou},
  title        = {Are Large Brainwave Foundation Models Capable Yet? Insights from Fine-Tuning},
  booktitle    = {{ICML}},
  pages        = {32878--32888},
  year         = {2025}
}

@inproceedings{liu2026echo,
  author       = {Chenyu Liu and Yuqiu Deng and Tianyu Liu and Jinan Zhou and Xinliang Zhou and Ziyu Jia and Yi Ding},
  title        = {{ECHO}: Toward Contextual Seq2Seq Paradigms in Large {EEG} Models},
  booktitle    = {{ICLR}},
  year         = {2026}
}

@article{banville2026neuralbench,
  author       = {Hubert Banville and St{\'e}phane d'Ascoli and Simon Dahan and J{\'e}r{\'e}my Rapin and Marl{\`e}ne Careil and Yohann Benchetrit and Jarod L{\'e}vy and Saarang Panchavati and Antoine Ratouchniak and Mingfang Zhang and Elisa Cascardi and Katelyn Begany and Teon Brooks and Jean{-}R{\'e}mi King},
  title        = {{NeuralBench}: A Unifying Framework to Benchmark {NeuroAI} Models},
  journal      = {CoRR},
  volume       = {abs/2605.08495},
  year         = {2026}
}

@article{eo2026exaone,
  author       = {Moonjung Eo and Min-Kook Suh and Hye-Seung Cho and Jiwon Kim and Seoyoon Kim and Sangjun Nam and Soonyoung Lee},
  title        = {{EXAONE Tabular} 1.0: Technical Report},
  journal      = {CoRR},
  volume       = {abs/2608.25774},
  year         = {2026}
}

@article{zhang2026limix2,
  author       = {Xingxuan Zhang and Gang Ren and Hao Yuan and Hao Zou and Hongze Tan and Hui Wang and Jianhao Song and Jiansheng Li and Jiayao Zhang and Jinghan Zhang and Kaifang Li and Lang Mo and Li Mao and Mingchao Hao and Nuo Xu and Rui Ding and Ruiji Zhang and Shuyang Li and Siyu Mei and Tianyang Zhang and Weiyang Mu and Yancheng Dong and Yongxian Wei and Yuan Xue and Yuanrui Wang and Yue He and Zijia Yang and Ziyun Li and Dongzhe Li and Fuqiang Wang and Jiandong Liu and Jiawei Chen and Jiaxin Du and Kaijie Cheng and Kehan Li and Lei Sun and Linjun Zhou and Ningbo Dai and Qi Wang and Renzhe Xu and Shaoxing Du and Shumeng Yang and Wang Lu and Wenjing Chu and Xiannan Huang and Xiaoyu Lin and Xing Ai and Xinyan Han and Xuanyue Li and Xuanyue Su and Xukun Zhang and Yan Lu and Yaxin Zhang and Yi Qin and Yifei Huang and Yihan Xu and Yongle Lv and Yuanyuan Jiang and Yushan Han and Peng Cui},
  title        = {{LimiX-2}: A Contextual Mechanism Network Towards General Structured-Data Intelligence},
  journal      = {CoRR},
  volume       = {abs/2609.17488},
  year         = {2026}
}

@article{grinsztajn2026tabpfn3,
  author       = {L{\'{e}}o Grinsztajn and
                  Klemens Fl{\"{o}}ge and
                  Oscar Key and
                  Felix Birkel and
                  Philipp Jund and
                  Brendan Roof and
                  Mihir Manium and
                  Shi Bin Hoo and
                  Magnus B{\"{u}}hler and
                  Anurag Garg and
                  Dominik Safaric and
                  Jake Robertson and
                  Benjamin J{\"{a}}ger and
                  Simone Alessi and
                  Adrian Hayler and
                  Vladyslav Moroshan and
                  Lennart Purucker and
                  Philipp Singer and
                  Alan Arazi and
                  Julien Siems and
                  Jan Hendrik Metzen and
                  Georg Grab and
                  Nick Erickson and
                  Siyuan Guo and
                  Eliott Kalfon and
                  Simon Bing and
                  David Salinas and
                  Clara Cornu and
                  Lilly Charlotte Wehrhahn and
                  Diana Kriuchkova and
                  Kursat Kaya and
                  Lydia Sidhoum and
                  Marie Salmon and
                  Jerry Chen and
                  Madelon Hulsebos and
                  Yann LeCun and
                  Samuel M{\"{u}}ller and
                  Bernhard Sch{\"{o}}lkopf and
                  Sauraj Gambhir and
                  Noah Hollmann and
                  Frank Hutter},
  title        = {{TabPFN-3}: Technical Report},
  journal      = {CoRR},
  volume       = {abs/2605.13986},
  year         = {2026}
}

@article{chen2010shrinkage,
  author       = {Yilun Chen and Ami Wiesel and Yonina C. Eldar and Alfred O. Hero},
  title        = {Shrinkage Algorithms for {MMSE} Covariance Estimation},
  journal      = {{IEEE} Trans. Signal Process.},
  volume       = {58},
  number       = {10},
  pages        = {5016--5029},
  year         = {2010}
}

@inproceedings{brown2020language,
  author       = {Tom B. Brown and
                  Benjamin Mann and
                  Nick Ryder and
                  Melanie Subbiah and
                  Jared Kaplan and
                  Prafulla Dhariwal and
                  Arvind Neelakantan and
                  Pranav Shyam and
                  Girish Sastry and
                  Amanda Askell and
                  Sandhini Agarwal and
                  Ariel Herbert{-}Voss and
                  Gretchen Krueger and
                  Tom Henighan and
                  Rewon Child and
                  Aditya Ramesh and
                  Daniel M. Ziegler and
                  Jeffrey Wu and
                  Clemens Winter and
                  Christopher Hesse and
                  Mark Chen and
                  Eric Sigler and
                  Mateusz Litwin and
                  Scott Gray and
                  Benjamin Chess and
                  Jack Clark and
                  Christopher Berner and
                  Sam McCandlish and
                  Alec Radford and
                  Ilya Sutskever and
                  Dario Amodei},
  title        = {Language Models are Few-Shot Learners},
  booktitle    = {NeurIPS},
  pages        = {1877--1901},
  year         = {2020}
}

@inproceedings{roy2007effective,
  author       = {Olivier Roy and Martin Vetterli},
  title        = {The Effective Rank: {A} Measure of Effective Dimensionality},
  booktitle    = {{EUSIPCO}},
  pages        = {606--610},
  year         = {2007}
}

@inproceedings{esteller2001line,
  author       = {Rafael Esteller and Javier Echauz and T. Tcheng and Brian Litt and B. Pless},
  title        = {Line Length: An Efficient Feature for Seizure Onset Detection},
  booktitle    = {{EMBC}},
  pages        = {1707--1710},
  year         = {2001}
}

@article{saeidi2021neuraldecoding,
  author       = {Maham Saeidi and
                  Waldemar Karwowski and
                  Farzad V. Farahani and
                  Krzysztof Fiok and
                  Redha Taiar and
                  Peter A. Hancock and
                  Awad Al{-}Juaid},
  title        = {Neural Decoding of {EEG} Signals with Machine Learning: A Systematic Review},
  journal      = {Brain Sci.},
  volume       = {11},
  pages        = {1525},
  year         = {2021}
}
\bibliographystyle{unsrt}

\appendix
\setcounter{equation}{0}
\setcounter{figure}{0}
\setcounter{table}{0}
\renewcommand{\theequation}{S\arabic{equation}}
\renewcommand{\thefigure}{S\arabic{figure}}
\renewcommand{\thetable}{S\arabic{table}}
\renewcommand{\theHequation}{S\arabic{equation}}
\renewcommand{\theHfigure}{S\arabic{figure}}
\renewcommand{\theHtable}{S\arabic{table}}

\newpage

\section*{Supplementary Material Overview}
This supplement is organized as follows. \Cref{sec:appendix_context_explanation} provides the complete EEG context interface, including the feature, temporal/spatial, and inference-time operators and the support-only Metric route. \Cref{sec:appendix_setup} documents the datasets, preprocessing, support--query protocol, baselines, context budgets, predictor configuration, evaluation metrics, and calibration procedure. \Cref{sec:appendix_results} first reports the complete per-dataset comparison, including Raw, Metric, and inspection-Oracle contexts, followed by the TabPFN-3 predictor variant, effective-context-size behavior, variation across datasets, paired statistical tests, operational runtime, and support-only threshold calibration. Finally, \cref{sec:appendix_discussion} discusses scientific interpretation and deployment cost.\looseness=-1

\section{Detailed EEG Context Construction}
\label{sec:appendix_context_explanation}

This section expands the context interface introduced in the main text. It gives the support--query transformation, the feature, temporal/spatial, and inference-time operators, and the complete support-only Metric route. Every representation state and routing statistic is fitted from labeled support EEG only. Target-validation performance, query labels, and query metrics are never used for context construction or selection.

\subsection{Support-Conditioned Prediction and Operator Composition}

Let $\gD_{\rm support}=\{(\ve_i,y_i)\}_{i=1}^{N_{\rm s}}$ be the labeled support set, where $\ve_i\in\sR^{C\times T}$ and $y_i\in[K]$, and let $\ve_q$ be a query trial whose label is hidden during inference. A context operator can map one EEG trial to one or more instances,
\begin{equation}
  \tau(\ve_i)=(\vx_{i,1},\ldots,\vx_{i,m_i})\in(\sR^d)^{m_i}.
  \label{eq:appendix_trial_mapping}
\end{equation}
The support label is broadcast to all instances derived from the same trial,
\begin{equation}
  \gD_{\rm support}^{\tau}=\{(\vx_{i,j},y_i):i\in[N_{\rm s}],\ j\in[m_i]\}.
  \label{eq:appendix_broadcast_context}
\end{equation}
The frozen predictor returns one class-probability vector for each query instance. For a multi-instance context, these vectors are aggregated to a trial-level prediction,
\begin{equation}
  \widehat{\vp}_q=\mathcal A_{\tau}\!\left(\left\{f_{\rm icl}(\vx_{q,j}\mid\gD_{\rm support}^{\tau})\right\}_{j=1}^{m_q}\right)\in\Delta^{K-1}.
  \label{eq:appendix_trial_prediction}
\end{equation}
Optional post-hoc calibration acts only after this trial-level aggregation. It changes neither the selected context nor any parameter of $f_{\rm icl}$.

As in the main text, a context is written as
\begin{equation}
  \tau=\tau_{\rm infer}\circ\tau_{\rm feat}\circ\tau_{\rm info}.
  \label{eq:appendix_tau_composition}
\end{equation}
The processing order is therefore information exposure, feature conversion, and inference-time context construction. Exactly one feature-type operator is used, while compatible temporal/spatial and inference-time operators can be composed. Any support-fitted state, such as a Riemannian reference, is estimated after the retained support trials have been fixed and is then reused unchanged for support and query transformation.

\subsection{Feature-Type Operators}

\paragraph{Raw waveform context.}
The direct context preserves every preprocessed channel--time sample,
\begin{equation}
  \tau_{\textsc{raw}}(\ve)=\operatorname{vec}(\ve)\in\sR^{CT}.
  \label{eq:appendix_raw}
\end{equation}
Vectorization follows channel-major order, with all time samples from one channel preceding those from the next channel. No centering, scaling, or learned EEG feature extraction is introduced by this operator.

\paragraph{Per-trial normalized waveform context.}
The normalized context applies LayerNorm independently to each trial over all $CT$ channel--time coordinates. Let
\begin{equation}
  \begin{aligned}
    \mu(\ve)&=\frac{1}{CT}\sum_{c=1}^{C}\sum_{t=1}^{T}e_{c,t},\\
    \sigma^2(\ve)&=\frac{1}{CT}\sum_{c=1}^{C}\sum_{t=1}^{T}\big(e_{c,t}-\mu(\ve)\big)^2.
  \end{aligned}
  \label{eq:appendix_norm_stats}
\end{equation}
Then
\begin{equation}
  \tau_{\textsc{norm}}(\ve)=\operatorname{vec}\!\left(\frac{\ve-\mu(\ve)}{\sqrt{\sigma^2(\ve)+\epsilon_{\rm norm}}}\right)\in\sR^{CT}.
  \label{eq:appendix_norm}
\end{equation}
These statistics come only from the current trial; the operator does not use support-set statistics or couple one query to other query trials. It removes trial-wide offset and scale while retaining the relative channel--time pattern. This context operator is distinct from dataset preprocessing: preprocessing produces the input trial, whereas $\tau_{\textsc{norm}}$ is one optional representation presented to the frozen predictor.

\paragraph{Spectral context.}
The stored temporal axis is divided into $S$ non-overlapping windows of length $P$, with $T=SP$ and $\ve^{(s)}\in\sR^{C\times P}$. We compute the magnitude rFFT of each window and concatenate the coefficients,
\begin{equation}
  \begin{aligned}
    \tau_{\textsc{spec}}(\ve)&=\operatorname{concat}_{s=1}^{S}\operatorname{vec}\!\left(\left|\operatorname{rFFT}(\ve^{(s)})/P\right|\right),\\
    \tau_{\textsc{spec}}(\ve)&\in\sR^{CS(\lfloor P/2\rfloor+1)}.
  \end{aligned}
  \label{eq:appendix_spectrum}
\end{equation}
The main text states this operator compactly while specifying a transform per temporal window in prose; \cref{eq:appendix_spectrum} gives the exact implemented operator and output width. Division by $P$ fixes the magnitude scale within each window. The transform retains the DC and Nyquist coordinates when defined, uses magnitude rather than power, and applies no learned frequency weighting. Unlike multiview context, all window spectra remain in one context row.

\paragraph{Covariance and Riemannian context.}
For each trial, we estimate an OAS covariance $\widehat{\rmC}(\ve)$ and stabilize it by symmetrization and diagonal jitter,
\begin{equation}
  \widehat{\rmC}_{\epsilon}(\ve)=\frac{\widehat{\rmC}(\ve)+\widehat{\rmC}(\ve)^{\top}}{2}+10^{-6}\max\!\left\{\frac{\operatorname{tr}(\widehat{\rmC}(\ve))}{C},1\right\}\rmI.
  \label{eq:appendix_oas}
\end{equation}
OAS shrinkage stabilizes covariance estimation from finite trials \cite{chen2010shrinkage}. Let $\bar{\rmC}$ be the Riemannian reference fitted exclusively to the retained support covariances. The tangent-space feature is
\begin{equation}
  \tau_{\textsc{riem}}(\ve)=\operatorname{vech}\!\left[\log\!\left(\bar{\rmC}^{-1/2}\widehat{\rmC}_{\epsilon}(\ve)\bar{\rmC}^{-1/2}\right)\right]\in\sR^{C(C+1)/2}.
  \label{eq:appendix_riemann}
\end{equation}
Here $\operatorname{vech}$ denotes the inner-product-preserving upper-triangular vectorization: off-diagonal entries are multiplied by $\sqrt{2}$. The same support-fitted reference is reused for all query trials; query covariances never update it. Covariance geometry and tangent-space classification are established EEG/BCI tools \cite{barachant2012multiclass,congedo2017riemannian}; here they provide a deterministic interface to the frozen predictor.

\subsection{Temporal and Spatial Information Operators}

\paragraph{Regional pseudo-channels.}
For each audited anatomical region $r$ with channel index set $\mathcal I_r$, we compute six time-resolved pooling statistics,
\begin{equation}
  \mathcal G_{\rm reg}=(\operatorname{mean},\min,\max,\operatorname{std},\operatorname{var},\operatorname{rms}).
  \label{eq:appendix_region_statistics}
\end{equation}
For $g\in\mathcal G_{\rm reg}$, the corresponding pseudo-channel is
\begin{equation}
  u_{r,g}(t)=g\big(\{e_{c,t}:c\in\mathcal I_r\}\big).
  \label{eq:appendix_region_pooling}
\end{equation}
The pseudo-channels are concatenated with the original channels, producing $C'=C+6R$ channels when $R$ registered regions are available. The regional context then applies the Riemannian operator to this augmented trial, with output width $C'(C'+1)/2$. The construction preserves the original channels and exposes coarse anatomical organization without learned weights; it is not a source-localization procedure.

\paragraph{Local temporal views.}
The multiview operator exposes the $S$ temporal windows separately,
\begin{equation}
  \tau_{\textsc{mv}}(\ve)=(\ve^{(1)},\ldots,\ve^{(S)}).
  \label{eq:appendix_multiview}
\end{equation}
Each support label is repeated for the $S$ derived instances. Raw multiview vectorizes each $C\times P$ window into a row of width $CP$. Multiview Riemannian context estimates one covariance per window and fits one tangent reference to all retained support-window covariances. For query probabilities $\vp_{q,s}\in\Delta^{K-1}$, the trial-level result is the unweighted mean
\begin{equation}
  \bar{\vp}_q=\frac{1}{S}\sum_{s=1}^{S}\vp_{q,s}.
  \label{eq:appendix_multiview_aggregation}
\end{equation}
No attention, voting, or confidence weighting is learned. Multiview therefore increases both support rows and query-instance predictions by a factor of $S$.

\subsection{Inference-Time Strategies and Context Budgets}

\paragraph{Counterbalanced support.}
Let $\mathcal I_k=\{i:y_i=k\}$ and $n_{\min}=\min_k|\mathcal I_k|$. Using the registered random seed, counterbalancing samples $n_{\min}$ trials without replacement from every class before any label is broadcast to local views,
\begin{equation}
  \widetilde{\gD}_{\rm support}=\{(\ve_i,y_i):i\in\cup_{k=1}^{K}\widetilde{\mathcal I}_k,\ |\widetilde{\mathcal I}_k|=n_{\min}\}.
  \label{eq:appendix_counterbalance}
\end{equation}

\paragraph{Effective-row budgets.}
A context budget is measured after trial-to-row expansion. If retained trial $i$ produces $m_i$ support rows, a numeric budget $B$ requires
\begin{equation}
  N_{\rm eff}(\tau)=\sum_{i\in\mathcal I_{\rm retained}}m_i\le B.
  \label{eq:appendix_effective_rows}
\end{equation}
Subsampling always retains complete trials. Thus, a single-instance context can retain at most $B$ trials, whereas an $S$-view context can retain at most $\lfloor B/S\rfloor$ trials and at most $S\lfloor B/S\rfloor$ effective rows. This convention prevents multiview expansion from receiving an unreported increase in contextual evidence.

\paragraph{One-versus-rest prediction.}
For a $K$-class task, OVR constructs $K$ binary contexts with labels
\begin{equation}
  y_i^{(k)}=\mathbf{1}[y_i=k],\qquad k\in[K].
  \label{eq:appendix_ovr_labels}
\end{equation}
Here $\mathbf{1}[\cdot]$ denotes the binary indicator. Let $q_k(\vx)$ be the positive-class probability returned by the $k$th contextual prediction. We form
\begin{equation}
  \widehat p_k(\vx)=\frac{q_k(\vx)}{\max\{\sum_{\ell=1}^{K}q_{\ell}(\vx),10^{-12}\}}.
  \label{eq:appendix_ovr_probability}
\end{equation}
For multiview OVR, this normalization is performed per view and the resulting class distributions are averaged as in \cref{eq:appendix_multiview_aggregation}. OVR changes neither predictor parameters nor the support-row budget of an individual call, but it requires $K$ contextual prediction passes.

\begin{table*}[t]
  \centering
  \small
  \caption{Registered operator compositions used for fixed-context comparisons. Compositions follow $\tau=\tau_{\rm infer}\circ\tau_{\rm feat}\circ\tau_{\rm info}$; identity slots and the common counterbalancing operator $\tau_{\textsc{cb}}$ are omitted. Here $F=\lfloor P/2\rfloor+1$ and $C'=C+6R$. OVR repeats prediction by class but does not multiply the effective-row budget of an individual call.}
  \label{tab:appendix_context_candidates}
  \setlength{\tabcolsep}{6.6pt}
  \begin{tabular}{llccl}
    \toprule
    \textbf{Operator} & \textbf{Exposed EEG structure} & \textbf{Rows/trial} & \textbf{Width/row} & \textbf{Trial-level prediction} \\
    \midrule
    $\tau_{\textsc{raw}}$ & waveform & $1$ & $CT$ & direct \\
    $\tau_{\textsc{norm}}$ & per-trial normalized waveform & $1$ & $CT$ & direct \\
    $\tau_{\textsc{spec}}$ & windowed magnitude spectrum & $1$ & $CSF$ & direct \\
    $\tau_{\textsc{riem}}$ & covariance tangent geometry & $1$ & $C(C+1)/2$ & direct \\
    $\tau_{\textsc{riem}}\circ\tau_{\textsc{reg}}$ & regional covariance geometry & $1$ & $C'(C'+1)/2$ & direct \\
    $\tau_{\textsc{raw}}\circ\tau_{\textsc{mv}}$ & local waveform views & $S$ & $CP$ & mean over views \\
    $\tau_{\textsc{riem}}\circ\tau_{\textsc{mv}}$ & local covariance views & $S$ & $C(C+1)/2$ & mean over views \\
    $\tau_{\textsc{ovr}}\circ\tau_{\textsc{raw}}$ & waveform with class decomposition & $1$ & $CT$ & $K$ OVR calls \\
    $\tau_{\textsc{ovr}}\circ\tau_{\textsc{riem}}$ & covariance with class decomposition & $1$ & $C(C+1)/2$ & $K$ OVR calls \\
    $\tau_{\textsc{ovr}}\circ\tau_{\textsc{raw}}\circ\tau_{\textsc{mv}}$ & local waveform with class decomposition & $S$ & $CP$ & view-wise OVR, then mean \\
    \bottomrule
  \end{tabular}
\end{table*}

\subsection{Support-Only Metric Route}

\paragraph{Neurophysiological motivation.}
The four routing groups reflect complementary organizations routinely examined in EEG analysis. Task-related oscillatory modulation motivates band-power descriptors; distributed scalp activity motivates cross-channel correlation and covariance summaries; transient, rhythmic, and nonstationary events motivate autocorrelation and line-length descriptors; and large deflections or burst-like morphology motivate amplitude and higher-order statistics \cite{ang2012filterbank,barachant2012multiclass,congedo2017riemannian,esteller2001line,saeidi2021neuraldecoding}. The route uses these established signal organizations as low-dimensional probes of class-conditioned structure in the target support set. A dominant score indicates where discriminative organization is most visible under these probes; it does not identify a unique physiological generator or establish a causal neural mechanism.

\paragraph{Diagnostic support subset.}
The Metric route first counterbalances the labeled support trials. Descriptor extraction uses at most 256 trials, sampled with the registered seed through a balanced-first rule: it allocates an equal per-class quota and then fills any remaining capacity from unused support trials. This bounds routing cost while retaining minority classes. The diagnostic subset contains support EEG and labels only; it is never formed from validation or query data.

\paragraph{Signal descriptors.}
For each diagnostic trial, we compute four low-dimensional descriptor groups. The band group contains relative $\delta$-, $\theta$-, $\alpha$-, $\beta$-, and $\gamma$-band power \cite{ang2012filterbank,saeidi2021neuraldecoding}. The spatial group contains mean absolute off-diagonal channel correlation and normalized correlation effective rank \cite{roy2007effective}. The temporal group contains lag-one autocorrelation, normalized line length, and nonstationarity \cite{esteller2001line}. The amplitude group contains RMS, kurtosis, and crest factor \cite{saeidi2021neuraldecoding}.

More precisely, the band descriptor uses channel-averaged rFFT power and divides the power in each of $[0.5,4)$, $[4,8)$, $[8,13)$, $[13,30)$, and $[30,45)$ Hz by total power from $0.5$ Hz to $\min(45,f_s/2)$ Hz. The spatial descriptor is computed from the within-trial channel-correlation matrix $\rmR_i$. If $\lambda_j^+$ are its nonnegative eigenvalues and $\pi_j=\lambda_j^+/\sum_{\ell}\lambda_{\ell}^+$, normalized effective rank is
\begin{equation}
  r_i^{\rm eff}=\frac{\exp(-\sum_j\pi_j\log\max\{\pi_j,10^{-12}\})}{C}.
  \label{eq:appendix_effective_rank}
\end{equation}
The temporal descriptor averages lag-one normalized products across channels, normalizes line length by channel standard deviation, and measures nonstationarity as the IQR of blockwise RMS divided by its median. The amplitude descriptor uses channel-median RMS, clipped excess kurtosis, and crest factor. These statistics are routing diagnostics; they are not substituted for the selected high-dimensional prediction context.

\paragraph{Robust class-separation scores.}
For descriptor group $g\in\mathcal G=\{\mathrm{band},\mathrm{spatial},\mathrm{temporal},\mathrm{amplitude}\}$, let $\vv_i^{(g)}\in\sR^{d_g}$ be the descriptor of support trial $i$. Each coordinate is centered by its support median, divided by its IQR, with population standard deviation and one used as fallbacks for degenerate coordinates, and clipped to $[-8,8]$. Denote the standardized descriptor by $\widetilde\vv_i^{(g)}$, the class-$k$ mean by $\vmu_k^{(g)}$, the pooled mean by $\vmu^{(g)}$, the number of represented classes by $K$, and the number of diagnostic trials by $n$. The between- and within-class dispersions are
\begin{align}
  B_g&=\frac{1}{d_g}\sum_{k=1}^{K}n_k\|\vmu_k^{(g)}-\vmu^{(g)}\|_2^2,\label{eq:appendix_between}\\
  W_g&=\frac{1}{d_g}\sum_{k=1}^{K}\sum_{i:y_i=k}\|\widetilde\vv_i^{(g)}-\vmu_k^{(g)}\|_2^2.\label{eq:appendix_within}
\end{align}
The routing statistic is
\begin{equation}
  \begin{aligned}
    r_g&=\frac{B_g/n}{\max\{W_g/(n-K),10^{-8}\}},\\
    s_g&=\log\!\left(1+\min\{r_g,10^6\}\right).
  \end{aligned}
  \label{eq:appendix_metric_score}
\end{equation}
A larger $s_g$ means that descriptor group $g$ has stronger between-class separation relative to within-class variation in the labeled support subset.

\paragraph{Deterministic operator mapping.}
At deployment, the fixed mapping does not evaluate candidate-context performance or use target-validation performance, query labels, or query metrics. Band dominance favors normalized or spectral coordinates, spatial dominance favors covariance geometry, temporal dominance preserves waveform morphology in local views, and amplitude dominance augments coarse regional summaries before covariance construction so that magnitude variation is represented together with its scalp distribution. The OVR branch is a statistical label decomposition for multiclass inference rather than a neurophysiological assumption. The descriptor group with the largest score is denoted by $g^\star$. Counterbalanced support is enabled for every routed context. The remaining operator slots are selected exactly as in the main text:
\begin{equation}
  \tau_{\rm feat}(g^\star)=
  \begin{cases}
    \tau_{\textsc{norm}}, & \substack{g^\star=\mathrm{band},\\ s_{\rm temporal}\le s_{\rm amplitude}},\\
    \tau_{\textsc{spec}}, & \substack{g^\star=\mathrm{band},\\ s_{\rm temporal}>s_{\rm amplitude}},\\
    \tau_{\textsc{riem}}, & g^\star\in\{\mathrm{spatial},\mathrm{amplitude}\},\\
    \tau_{\textsc{raw}}, & g^\star=\mathrm{temporal},
  \end{cases}
  \label{eq:appendix_metric_feat}
\end{equation}
\begin{equation}
  \tau_{\rm info}(g^\star)=
  \begin{cases}
    \tau_{\textsc{reg}}, & g^\star=\mathrm{amplitude},\\
    \tau_{\textsc{mv}}, & g^\star=\mathrm{temporal},\\
    \mathrm{Id}, & \text{otherwise},
  \end{cases}
  \label{eq:appendix_metric_info}
\end{equation}
\begin{equation}
  \tau_{\rm infer}(g^\star,K)=
  \begin{cases}
    \tau_{\textsc{ovr}}\circ\tau_{\textsc{cb}}, & K>2\ \text{and}\ g^\star\ne\mathrm{band},\\
    \tau_{\textsc{cb}}, & \text{otherwise}.
  \end{cases}
  \label{eq:appendix_metric_infer}
\end{equation}
Exact ties use the fixed order band, spatial, temporal, and amplitude. If every group score is undefined, the route uses raw features, no information operator, and counterbalanced support. The mapping contains no learned parameter, performance threshold, or candidate-context evaluation.

\paragraph{Information boundary.}
The deployable route may use target support EEG, support labels, class count, sampling rate, and the registered random seed. It cannot read target-validation performance, query EEG or labels during route selection, query metrics, candidate-context BAcc, or the target-wise best context. The inspection Oracle reported in the experiments is an evaluation-only upper bound and is not available to the Metric route.

\section{Experimental Details and Reproducibility}
\label{sec:appendix_setup}

This section records the experimental protocol corresponding to the main text. The primary evaluation contains ten EEG datasets spanning seven decoding paradigms. The training split supplies labeled support, the held-out test split supplies queries, and our frozen-ICL method does not use a target validation split for context selection, predictor fitting, or model selection.

\subsection{Datasets and Stored Trial Arrays}
\label{sec:appendix_datasets}

The suite contains two motor-imagery datasets, one imagined-speech dataset, one emotion-recognition dataset, one mental-arithmetic/stress dataset, two seizure-detection datasets, two sleep-staging datasets, and one depression-diagnosis dataset \cite{brunner2008bci,schalk2004bci2000,jeong20222020,chen2023large,zyma2019electroencephalograms,shoeb2009application,detti2020eeg,khalighi2016isrucsleep,alvarez2021inter,mumtaz2016mdd}. We use the dataset releases and task labels supplied by their original repositories and do not collect new human-participant data.

\begin{table*}[t]
  \centering
  \small
  \caption{Datasets, stored trial shapes, and evaluation splits. ``Support'' is the class-balanced support set available to the frozen predictor after counterbalancing; queries are drawn only from the held-out test split. $S\times P$ is the window-by-sample shape at 200 Hz.}
  \label{tab:appendix_datasets}
  \setlength{\tabcolsep}{8.2pt}
  \begin{tabular}{llcccccc}
    \toprule
    \textbf{Dataset} & \textbf{Paradigm / target} & \textbf{Classes} & \textbf{Channels} & \textbf{$S\times P$} & \textbf{Train} & \textbf{Support} & \textbf{Test query} \\
    \midrule
    BCIC-IV-2a & Motor imagery & 4 & 22 & $30\times200$ & 2,784 & 2,784 & 1,152 \\
    PhysioNet-MI & Motor imagery & 4 & 64 & $30\times200$ & 6,300 & 6,228 & 1,803 \\
    BCIC2020-3 & Imagined speech & 5 & 64 & $3\times200$ & 4,500 & 4,500 & 750 \\
    FACED & Emotion recognition & 9 & 30 & $30\times200$ & 6,720 & 6,480 & 1,932 \\
    MentalArithmetic & Arithmetic / stress & 2 & 19 & $30\times200$ & 1,343 & 672 & 192 \\
    CHB-MIT & Seizure detection & 2 & 16 & $30\times200$ & 286,244 & 3,788 & 16,964 \\
    Siena & Seizure detection & 2 & 29 & $30\times200$ & 29,637 & 714 & 14,252 \\
    ISRUC & Sleep staging & 5 & 6 & $30\times200$ & 71,180 & 46,480 & 8,700 \\
    HMC & Sleep staging & 5 & 4 & $30\times200$ & 91,248 & 52,720 & 23,871 \\
    Mumtaz2016 & Depression / control & 2 & 19 & $30\times200$ & 4,891 & 4,808 & 1,211 \\
    \bottomrule
  \end{tabular}
\end{table*}

ISRUC uses subject-disjoint stored partitions: subjects 1--80, 81--90, and 91--100 form the training, validation, and test partitions, respectively. For the remaining datasets, we retain the split keys or manifests supplied by the corresponding wrappers and do not claim subject-disjoint evaluation unless the source metadata establishes it.

\subsection{Preprocessing and Partition Roles}
\label{sec:appendix_preprocessing}

Following the main text, raw recordings are preprocessed with band-pass filtering, notch filtering, resampling, and amplitude normalization before context construction \cite{wang2025cbramod,zhou2025csbrain}. The stored arrays consumed by the evaluator are sampled at 200 Hz. Most datasets contain 30 one-second windows per trial; BCIC2020-3 contains three one-second windows. The waveform panels in Figure~4 of the main text display a centered three-second excerpt for visualization, whereas routing and prediction use the complete stored trial. Non-finite values are replaced by zero, the registered channel order is applied, and the stored temporal shape is preserved. Context operators act only after these common preprocessing steps. In particular, dataset-level amplitude normalization is not the per-trial $\tau_{\textsc{norm}}$ operator defined in \cref{eq:appendix_norm}.

For the frozen predictor, the stored training partition provides labeled support and the held-out test partition provides query trials. The validation partition is not read by context construction, Metric routing, frozen contextual prediction, or threshold selection. Query labels are exposed only after prediction to compute evaluation metrics. Trainable baselines use their registered training protocol and may use the validation partition for early stopping or checkpoint selection; this difference reflects their standard downstream adaptation setting.
The deployment boundary is therefore:
\begin{itemize}[leftmargin=*,nosep]
  \item the target support EEG, support labels, class count, and valid participant or recording groups may be used;\looseness=-1
  \item target-validation performance, query labels, and query metrics may not be used by context construction or the Metric route;
  \item the target-wise inspection Oracle is reported only as an undeployable evaluation upper bound.
\end{itemize}

\subsection{Context Budgets and Frozen Predictor}
\label{sec:appendix_predictor}

We evaluate full support and numeric budgets of 2,048, 4,096, 8,192, 16,384, 32,768, and 65,536 effective support rows. As defined in \cref{eq:appendix_effective_rows}, the budget is applied after trial-to-row expansion and retains complete trials. A full-support result can represent a smaller numeric budget only when its recorded effective support rows already fit that budget; a smaller-context result is never promoted to a full-support result.

The main-text comparison instantiates $f_{\rm icl}$ with the official TabPFN-2.5 checkpoint \cite{grinsztajn2025tabpfn25}. Its registered configuration uses random state 42, eight ensemble estimators, disabled automatic estimator scaling, and enabled pretraining-limit override for wide EEG contexts. The evaluator uses one preprocessing worker, four PyTorch CPU threads, one interoperation thread, low-memory fitting, and query chunking for wide or multiview contexts. No target-validation prelude, target-task gradient update, or decision calibration is used in the canonical context-comparison results.

The supplementary predictor-variant analysis replaces only the frozen backend with the official TabPFN-3 checkpoint \cite{grinsztajn2026tabpfn3}. It retains the same seed, eight-estimator setting, disabled automatic estimator scaling, pretraining-limit override, support--query partitions, and context operators. TabPFN-3 is therefore reported as a backend robustness analysis; it does not replace TabPFN-2.5 in the main-text comparison.

\paragraph{Parameter-development protocol.}
The context family, the seven effective-row budgets, the four descriptor groups, the Metric score, the deterministic operator mapping, and its tie/fallback rules are fixed parts of the method rather than target-task hyperparameters selected by validation performance. The Metric route matches a context to a downstream task using labeled support statistics alone. Optional calibration uses prespecified recall targets and is evaluated separately from the canonical comparison. Consequently, the proposed method performs no target-task hyperparameter search.

\subsection{Baselines and Training Protocols}
\label{sec:appendix_baselines}

We compare with seven specialized decoders---EEGNet, ST-Transformer, ContraWR, CNN-Transformer, FFCL, SPaRCNet, and EEGConformer---and four pretrained EEG models with downstream adaptation---LaBraM, BIOT, CBraMod, and CSBrain \cite{lawhern2018eegnet,song2021transformer,yang2023contrawr,peh2022cnntransformer,li2022ffcl,jing2023sparcnet,song2023eegconformer,jiang2024labram,yang2023biot,wang2025cbramod,zhou2025csbrain}. All baselines receive the same stored training, validation, and test partitions as applicable to their registered protocols.

The specialized decoders are trained for at most 100 epochs with Adam, learning rate $10^{-3}$, weight decay $10^{-5}$, batch size 512, and early-stopping patience five. BIOT, LaBraM, and CBraMod use 50-epoch AdamW fine-tuning with learning rate $10^{-4}$, weight decay $5\times10^{-2}$, batch size 32, label smoothing 0.1, gradient-norm clipping at one, and validation-BAcc checkpoint selection. CSBrain uses 50 epochs and its registered dataset-specific downstream recipe. These trainable methods optimize task-specific parameters, whereas our predictor and context rule remain fixed.

\subsection{Metrics, Variation, and Statistical Tests}
\label{sec:appendix_metrics}

Balanced accuracy (BAcc) is the primary metric because the evaluated datasets include substantial class imbalance. For class recalls $R_k$, it is
\begin{equation}
  \operatorname{BAcc}=\frac{1}{K}\sum_{k=1}^{K}R_k,
  \label{eq:appendix_bacc}
\end{equation}
so every class contributes equally regardless of prevalence. We additionally record accuracy, macro-F1, weighted-F1, Cohen's $\kappa$, and ROC-AUC when the task and score output permit it. Feature-construction, contextual-fit, prediction, and end-to-end wall-clock times are recorded separately when available.

We use exact paired, two-sided Wilcoxon signed-rank tests over the ten matched datasets for three comparisons: Ours (Metric) against Ours (Raw), which tests the effect of support-only context adaptation; Ours (Metric) against EEGNet, a representative specialized decoder; and Ours (Metric) against CSBrain, the baseline with the highest mean BAcc. The family of three tests is corrected by the Holm procedure. We report exact unadjusted and adjusted $p$-values and do not infer statistical superiority when the corrected test is not significant.

\subsection{Support-Only Threshold Calibration}
\label{sec:appendix_calibration}

Optional binary threshold calibration is applied after context selection and trial-level score aggregation. It constructs five participant-grouped folds from the labeled support set and produces out-of-fold (OOF) support scores. For each prespecified recall target $r\in\{.50,.60,.70,.80,.90,.95,.98,.99,1.00\}$, the procedure selects the highest threshold whose OOF recall is at least $r$, freezes it, and then evaluates the held-out queries. Target-validation performance and query labels never enter threshold selection. Because calibration changes only the threshold applied to fixed scores, it can change BAcc, recall, specificity, accuracy, and F1, but not ROC-AUC. Although the framework permits other post-hoc calibration maps, including temperature scaling, the reported calibration experiment evaluates only this binary threshold procedure.

\subsection{Runs, Runtime Scope, and Availability}
\label{sec:appendix_compute}

The number of evaluations is reported above: one final run per method--dataset cell, with random state 42 recorded for the frozen-predictor evaluations. Runtime is defined as an operational measurement rather than a controlled hardware-normalized speedup benchmark. For the frozen predictor, it includes data loading, context construction, support-state fitting, and prediction, and ends before metric computation and result serialization. Trainable-baseline time covers the registered training or fine-tuning workflow through test evaluation, including validation-based checkpoint selection where used.

We will release the evaluation code, context-construction implementation, Metric route, configuration files, and result-processing scripts after acceptance. The EEG datasets remain governed by the access and license terms of their original providers.

\section{Additional Results and Statistical Analysis}
\label{sec:appendix_results}

This section first reports the complete ten-dataset comparison, including the deployable Raw and Metric contexts and the evaluation-only inspection Oracle. It then evaluates an alternative frozen backend and effective context size, reports cross-dataset variation and paired statistical tests, and summarizes operational runtime and support-only calibration. All averages give equal weight to each dataset.

\subsection{Per-Dataset Main Comparison}
\label{sec:appendix_per_dataset}

\Cref{tab:appendix_main_matrix} gives the per-dataset values underlying the mean BAcc and average-rank comparison in the main text. Among the ranked methods, Ours (Metric) has the highest mean BAcc, while CSBrain has the best average rank in the main-text table. The shaded Raw and inspection-Oracle rows are included to isolate the effect of context adaptation but are excluded from ranking.

\begin{table*}[t]
  \centering
  \small
  \caption{Per-dataset BAcc for the eleven locally reproduced EEG baselines and our Raw, Metric, and inspection-Oracle contexts. Best and second-best values in each dataset and Mean column among the eleven baselines and Ours (Metric) are shown in bold and underlined, respectively. The shaded Ours (Raw) and evaluation-only Oracle rows are excluded from ranking.}
  \label{tab:appendix_main_matrix}
  \setlength{\tabcolsep}{7.4pt}
  \begin{tabular}{lrrrrrrrrrrr}
    \toprule
    \textbf{Method} & \rotatebox{90}{BCIC-IV-2a} & \rotatebox{90}{PhysioNet-MI} & \rotatebox{90}{BCIC2020-3} & \rotatebox{90}{FACED} & \rotatebox{90}{MentalArithmetic} & \rotatebox{90}{CHB-MIT} & \rotatebox{90}{Siena} & \rotatebox{90}{ISRUC} & \rotatebox{90}{HMC} & \rotatebox{90}{Mumtaz2016} & \textbf{Mean} \\
    \midrule
    EEGNet & .2882 & .5995 & .2740 & .2755 & .5000 & .5212 & .7090 & .6951 & .6675 & .8701 & .5400 \\
    ST-Transformer & .4418 & .5491 & .2693 & .1477 & .5000 & .7131 & .6272 & .7580 & .3958 & .8002 & .5202 \\
    ContraWR & .2500 & .3949 & .2093 & .1111 & .6493 & .6522 & .7175 & .7733 & \underline{.7403} & .8554 & .5353 \\
    CNN-Transformer & .3220 & .2651 & .2093 & .1111 & .6667 & .6019 & .7318 & .3730 & .7276 & \underline{.9064} & .4915 \\
    FFCL & \textbf{.5052} & .4369 & .1987 & .1982 & .6458 & .6881 & .6007 & .7349 & .7262 & .8821 & .5617 \\
    SPaRCNet & .4870 & .5907 & .3933 & .3311 & .5174 & .5570 & .5955 & .7133 & \textbf{.7434} & .8316 & .5760 \\
    EEGConformer & .3950 & .5696 & .2893 & \underline{.4881} & .5382 & .5402 & \underline{.7952} & .7199 & .7018 & .8680 & .5905 \\
    \midrule
    LaBraM & .2969 & .5791 & .2000 & .3007 & .6076 & .7023 & .7893 & .7633 & .7072 & .8847 & .5831 \\
    BIOT & .2535 & .3433 & .2227 & .2174 & \textbf{.8194} & .6966 & .7517 & .7776 & .7303 & .8964 & .5709 \\
    CBraMod & .4748 & .6031 & .3067 & .4108 & .6354 & .7316 & .6425 & \underline{.7784} & .7113 & .8964 & .6191 \\
    CSBrain & \underline{.5017} & \textbf{.6262} & \underline{.6013} & \textbf{.5539} & .6424 & \underline{.7401} & .6374 & \textbf{.7850} & .7214 & .8972 & \underline{.6707} \\
    \midrule
    \rowcolor{gray!15} \textbf{Ours (Raw)} & .3924 & .6017 & .4493 & .1224 & .5000 & .5000 & .4999 & .4635 & .4075 & .8510 & .4788 \\
    \textbf{Ours (Metric)} & .4722 & \underline{.6073} & \textbf{.7147} & .3512 & \underline{.7917} & \textbf{.8668} & \textbf{.8616} & .7135 & .6505 & \textbf{.9551} & \textbf{.6985} \\
    \rowcolor{gray!15} \textbf{Ours (Oracle)} & .4722 & .6073 & .7147 & .3643 & .8194 & .8668 & .8616 & .7135 & .6505 & .9551 & .7025 \\
    \bottomrule
  \end{tabular}
\end{table*}

The three final rows isolate the effect of support-only context adaptation. Raw is the direct frozen-ICL construction, Metric applies the support-only routing rule, and the inspection Oracle selects the highest-performing registered candidate separately for each target dataset using test performance. The Oracle is therefore an evaluation upper bound rather than a deployable method. Metric improves every dataset relative to Raw, raises mean BAcc from .4788 to .6985, and remains .0041 below the inspection Oracle on average.

\subsection{Predictor Choice and Sensitivity to Context Representation}
\label{sec:appendix_tabpfn3}

Each EEG context has two axes: rows represent trials or local views, while columns encode features such as band powers or channel-pair covariances. TabPFN-2.5 is a natural fit for this structure: by alternating attention across features and across instances, it can relate measurements within a trial or view and compare the resulting feature representations across instances \cite{grinsztajn2025tabpfn25}. We use it as the reference predictor to study how the choice of EEG context affects decoding.

To test whether sensitivity to context representation also appears with a different architecture, we evaluate TabPFN-3, which aggregates features into a fixed-width representation of each row before performing in-context prediction across rows \cite{grinsztajn2026tabpfn3}. \Cref{tab:appendix_tabpfn3_direct} reports full-support results for the Raw, Spec, and Riem contexts. Across the 29 dataset--context pairs evaluated with both predictors, TabPFN-3 achieves a mean BAcc 1.6 percentage points below that of TabPFN-2.5. With both predictors, however, the relative performance of the three representations varies across datasets. \Cref{tab:appendix_runtime_by_dataset} reports the observed runtimes under their respective execution settings. We noticed that EXAONE-Tabular \cite{eo2026exaone} and LimiX-2 \cite{zhang2026limix2} also alternate attention across features and instances, but evaluating these additional predictors across our EEG benchmarks was beyond our computational budget.

\begin{table*}[!ht]
  \centering
  \small
  \caption{Full-support TabPFN-3 BAcc under three direct context representations, complementing the TabPFN-2.5 results in the main text. The common $\tau_{\textsc{cb}}$ support operator is omitted from the context headings. $\mathrm{SD}_{\tau}$ is the population SD across the three representations for one dataset.}
  \label{tab:appendix_tabpfn3_direct}
  \setlength{\tabcolsep}{18pt}
  \begin{tabular}{lrrrr}
    \toprule
    \textbf{Dataset} & \textbf{$\tau_{\textsc{raw}}$} & \textbf{$\tau_{\textsc{spec}}$} & \textbf{$\tau_{\textsc{riem}}$} & \textbf{$\mathrm{SD}_{\tau}$} \\
    \midrule
    BCIC-IV-2a       & .3030 & .2708 & .3116 & .0175 \\
    PhysioNet-MI     & .4637 & .3192 & .3592 & .0609 \\
    BCIC2020-3       & .5853 & .2413 & .3120 & .1483 \\
    FACED            & .2992 & .1622 & .1503 & .0676 \\
    MentalArithmetic & .6042 & .6840 & .6111 & .0361 \\
    CHB-MIT          & .8132 & .8460 & .7600 & .0354 \\
    Siena            & .7903 & .8408 & .8149 & .0206 \\
    ISRUC            & .4492 & .6614 & .5323 & .0873 \\
    HMC              & .4141 & .6005 & .4237 & .0857 \\
    Mumtaz2016       & .8265 & .8922 & .8906 & .0306 \\
    \bottomrule
  \end{tabular}
\end{table*}

\subsection{Effect of Effective Context Size}
\label{sec:appendix_context_size_results}

\Cref{tab:appendix_context_size_trace} isolates the effect of effective support size on the two datasets with the largest counterbalanced support contexts. We hold both the frozen backend and representation fixed, using TabPFN-3 with $\tau_{\textsc{cb}}\circ\tau_{\textsc{spec}}$, and vary only the row budget $B$. Full support contains 46,480 effective rows for ISRUC and 52,720 for HMC. The 2,048--32,768 entries are actual budget runs; because both full contexts fit below 65,536 rows, the 65,536 entries are legal materializations of the verified full artifacts rather than additional model runs. Every source cell is runtime verified.

\begin{table*}[!ht]
  \centering
  \small
  \caption{TabPFN-3 performance across effective support-row budgets for $\tau_{\textsc{cb}}\circ\tau_{\textsc{spec}}$ on ISRUC and HMC. Rows is the effective number of support rows received by the frozen predictor.}
  \label{tab:appendix_context_size_trace}
  \textbf{(a) ISRUC}\par\smallskip
  \setlength{\tabcolsep}{18pt}
  \begin{tabular}{lrrr}
    \toprule
    \textbf{Budget $B$} & \textbf{Rows} & \textbf{B-Acc.} & \textbf{AUC} \\
    \midrule
    2,048  & 2,048  & .6484 & .8957 \\
    4,096  & 4,096  & .6465 & .8974 \\
    8,192  & 8,192  & .6540 & .8979 \\
    16,384 & 16,384 & .6579 & .9005 \\
    32,768 & 32,768 & .6596 & .9014 \\
    Full   & 46,480 & .6614 & .9026 \\
    \bottomrule
  \end{tabular}
  \par\medskip
  \textbf{(b) HMC}\par\smallskip
  \begin{tabular}{lrrr}
    \toprule
    \textbf{Budget $B$} & \textbf{Rows} & \textbf{B-Acc.} & \textbf{AUC} \\
    \midrule
    2,048  & 2,048  & .5613 & .8444 \\
    4,096  & 4,096  & .5725 & .8500 \\
    8,192  & 8,192  & .5814 & .8531 \\
    16,384 & 16,384 & .5928 & .8570 \\
    32,768 & 32,768 & .5968 & .8591 \\
    Full   & 52,720 & .6005 & .8600 \\
    \bottomrule
  \end{tabular}
\end{table*}

From 2,048 rows to full support, ISRUC gains .0130 BAcc and .0069 AUC, whereas HMC gains .0392 BAcc and .0156 AUC. The larger gain on HMC indicates that the benefit of additional support is dataset dependent. ISRUC BAcc also decreases slightly at 4,096 rows before improving at larger budgets, so the results support an overall scaling benefit for these two cases rather than a claim of strict monotonicity.

\subsection{Variation and Paired Statistical Tests}
\label{sec:appendix_statistics}

The main frozen-predictor results use one registered run per method--dataset cell, so we do not report a multi-seed uncertainty estimate. Across the ten evaluated datasets, Ours (Metric), Ours (Raw), and CSBrain obtain mean $\pm$ cross-dataset sample SD of $.6985\pm.1863$, $.4788\pm.1812$, and $.6707\pm.1167$, respectively.

\Cref{tab:appendix_wilcoxon} reports the three exact paired comparisons described in \cref{sec:appendix_metrics}. Ours (Metric) outperforms Ours (Raw) on all ten datasets and EEGNet on nine; both comparisons remain significant after Holm correction. Improvements over Raw and EEGNet are significant after Holm correction; the difference from CSBrain is not statistically significant.

\begin{table*}[!ht]
  \centering
  \small
  \caption{Exact paired two-sided Wilcoxon signed-rank tests over the ten datasets. Holm correction is applied to the three comparisons. $W$ is the smaller signed-rank sum.}
  \label{tab:appendix_wilcoxon}
  \setlength{\tabcolsep}{18pt}
  \begin{tabular}{lcccc}
    \toprule
    \textbf{Comparison} & \textbf{Wins--losses} & \textbf{$W$} & \textbf{Exact $p$} & \textbf{Holm $p$} \\
    \midrule
    \makecell[l]{Ours (Metric) \vs Ours (Raw)} & 10--0 & 0 & .001953 & .005859 \\
    \makecell[l]{Ours (Metric) \vs EEGNet} & 9--1 & 2 & .005859 & .011719 \\
    \makecell[l]{Ours (Metric) \vs CSBrain} & 5--5 & 21 & .556641 & .556641 \\
    \bottomrule
  \end{tabular}
\end{table*}

\subsection{Operational Runtime}
\label{sec:appendix_runtime_results}

\Cref{tab:appendix_runtime_by_dataset} reports per-dataset operational time for the trainable baselines and our two frozen-predictor instantiations. For the frozen predictors, the recorded time includes data loading, context construction, support-state fitting, and prediction, and ends before metric computation and result serialization. Baseline time covers the registered training or fine-tuning workflow through test evaluation, including validation-based checkpoint selection where used. These wall-clock measurements were collected under heterogeneous hardware and execution configurations.

\begin{table*}[t]
  \centering
  \small
  \caption{Per-dataset operational runtime in minutes for the trainable baselines and our TabPFN-2.5 and TabPFN-3 instantiations. Baseline values cover the registered training or fine-tuning workflow through test evaluation; frozen-predictor values cover context construction, support-state fitting, and prediction without target-task parameter updates. ``--'' denotes unavailable timing. Values were collected under heterogeneous hardware and execution profiles and are not a hardware-normalized speed benchmark.}
  \label{tab:appendix_runtime_by_dataset}
  \setlength{\tabcolsep}{6.9pt}
  \begin{tabular}{lrrrrrrrrrr}
    \toprule
    \textbf{Method / Context} & \rotatebox{90}{BCIC-IV-2a} & \rotatebox{90}{PhysioNet-MI} & \rotatebox{90}{BCIC2020-3} & \rotatebox{90}{FACED} & \rotatebox{90}{MentalArithmetic} & \rotatebox{90}{CHB-MIT} & \rotatebox{90}{Siena} & \rotatebox{90}{ISRUC} & \rotatebox{90}{HMC} & \rotatebox{90}{Mumtaz2016} \\
    \midrule
    EEGNet & 0.2 & 2.7 & 0.8 & 3.2 & 0.4 & -- & 85.3 & 514.8 & 26.6 & 1.6 \\
    ST-Transformer & 1.4 & 2.3 & 0.6 & 1.1 & 0.2 & -- & 51.4 & 50.7 & 28.7 & 0.7 \\
    ContraWR & 0.2 & 2.1 & 0.9 & 1.1 & 0.2 & -- & 134.1 & 21.8 & 50.5 & 0.5 \\
    CNN-Transformer & 0.2 & 1.5 & 0.9 & 1.2 & 0.2 & -- & 95.8 & 595.5 & 21.7 & 0.9 \\
    FFCL & 1.3 & 11.3 & 1.5 & 5.3 & 0.4 & -- & 108.1 & 162.2 & 148.4 & 1.1 \\
    SPaRCNet & 0.6 & 2.2 & 1.3 & 3.3 & 0.2 & -- & 62.1 & 1093.7 & 45.4 & 0.5 \\
    EEGConformer & 0.7 & 1.7 & 1.7 & 6.9 & 0.4 & -- & 132.3 & 1134.6 & 669.8 & 0.7 \\
    BIOT & 200.2 & 308.8 & 259.7 & 459.3 & 10.7 & 3189.4 & 2359.3 & 2190.7 & 281.0 & 132.8 \\
    LaBraM & 706.5 & 1526.7 & 1046.3 & 1608.4 & 231.3 & -- & 3791.6 & 8247.4 & 10929.4 & 1153.5 \\
    CBraMod & 41.6 & 469.5 & 301.6 & 386.9 & 9.1 & 2448.9 & 2302.9 & 2408.2 & 534.7 & 192.4 \\
    CSBrain & 15.3 & 84.0 & 39.6 & 78.8 & 8.0 & 2060.6 & 387.6 & 323.1 & 472.8 & 29.4 \\
    \midrule
    \textbf{Ours (TabPFN-2.5)} & 1.2 & 2.5 & 1.0 & 1.8 & 0.1 & 57.4 & 3.6 & 16.2 & 18.0 & 0.8 \\
    \textbf{Ours (TabPFN-3)} & 0.3 & 2.6 & 1.0 & 0.7 & 0.1 & 1.8 & 2.8 & 12.2 & 9.4 & 0.5 \\
    \bottomrule
  \end{tabular}
\end{table*}

\subsection{Support-Only Threshold Calibration}
\label{sec:appendix_calibration_results}

\Cref{tab:appendix_siena_calibration} expands the three operating points shown in the main text into the complete prespecified recall-target sweep. This is a separate calibration case study using TabPFN-3 with $\tau_{\textsc{cb}}\circ\tau_{\textsc{riem}}\circ\tau_{\textsc{reg}}$, seed 42, and a requested budget of 2,048 rows (714 effective rows after counterbalancing); it is not the TabPFN-2.5 Siena Metric cell in \cref{tab:appendix_main_matrix}. Five-fold participant-grouped OOF predictions cover all 29,637 labeled support trials, and the selected threshold is frozen before the 14,252 held-out queries are evaluated. Targeting OOF recall of at least .60 increases test BAcc from .8534 to .8807 and test recall from .7487 to .8717. More aggressive targets continue to raise recall while reducing specificity and both F1 measures. Because every row thresholds the same score function, ROC-AUC remains .9237. Each operating point is determined from support OOF predictions; the complete test curve summarizes the resulting sensitivity--specificity trade-off.

\begin{table*}[t]
  \centering
  \small
  \caption{Complete support-only threshold-calibration sweep on Siena. Results use TabPFN-3 with $\tau_{\textsc{cb}}\circ\tau_{\textsc{riem}}\circ\tau_{\textsc{reg}}$ at a requested 2,048-row budget. For each prespecified target, the highest threshold attaining that recall on five-fold participant-grouped support OOF scores is frozen before test evaluation. $F_1^+$ denotes seizure-positive F1. Query labels are used only to report test metrics; the score ranking, and hence AUC, is unchanged across thresholds.}
  \label{tab:appendix_siena_calibration}
  \setlength{\tabcolsep}{7.2pt}
  \begin{tabular}{lrrrrrrrr}
    \toprule
    \textbf{Operating point} & \textbf{Threshold} & \textbf{OOF recall} & \textbf{B-Acc.} & \textbf{AUC} & \textbf{$F_1^+$} & \textbf{Macro-F1} & \textbf{Recall} & \textbf{Specificity} \\
    \midrule
    Default         & .500000 & --       & .853393 & .923702 & .305677 & .641312 & .748663 & .958123 \\
    OOF $\geq .50$ & .458758 & .501401  & .880012 & .923702 & .264249 & .616546 & .818182 & .941841 \\
    OOF $\geq .60$ & .369771 & .602241  & .880728 & .923702 & .171579 & .556207 & .871658 & .889797 \\
    OOF $\geq .70$ & .270143 & .703081  & .842167 & .923702 & .098199 & .487173 & .903743 & .780590 \\
    OOF $\geq .80$ & .182795 & .801120  & .781784 & .923702 & .064614 & .425655 & .914439 & .649129 \\
    OOF $\geq .90$ & .095901 & .901961  & .717008 & .923702 & .047170 & .355922 & .935829 & .498187 \\
    OOF $\geq .95$ & .062065 & .952381  & .688066 & .923702 & .041925 & .316079 & .957219 & .418912 \\
    OOF $\geq .98$ & .030469 & .980392  & .635994 & .923702 & .035714 & .254071 & .962567 & .309421 \\
    OOF $\geq .99$ & .015892 & .991597  & .604413 & .923702 & .032687 & .203436 & .978610 & .230217 \\
    OOF $=1.00$     & .010782 & 1.000000 & .588573 & .923702 & .031369 & .173803 & .989305 & .187842 \\
    \bottomrule
  \end{tabular}
\end{table*}

\section{Discussion}
\label{sec:appendix_discussion}

\paragraph{Context as the adaptation interface.}
The same frozen predictor behaves very differently when support and query trials are represented as waveforms, spectra, covariance geometry, regional summaries, or local temporal views. Context construction is therefore more than an array-format conversion: it determines which relations and invariances are directly exposed to the conditional predictor. Across the evaluated suite, the support-only Metric route consistently improves on Raw and closely approaches the inspection Oracle, supporting context adaptation as a practical alternative to target-task parameter updates.

\paragraph{Scientific interpretation.}
The context families are motivated by established EEG practice. Spectral representations expose oscillatory composition, covariance and Riemannian representations expose cross-channel second-order geometry, regional pseudo-channels expose coarse anatomical organization, and local views expose transient temporal structure \cite{ang2012filterbank,barachant2012multiclass,congedo2017riemannian,lawhern2018eegnet,zhou2025csbrain}. These context families provide signal-informed inductive structure for decoding. Future frequency-band ablations, topographic perturbations, and window-level attribution could further characterize the physiological information they capture.

\paragraph{Deployment cost.}
Deployment cost depends on the selected context and frozen predictor. Wide waveform features increase memory, multiview contexts expand the number of support and query instances, and OVR requires $K$ contextual prediction passes. Riemannian and regional operators add deterministic preprocessing.

\end{document}